\documentclass[letterpaper]{article}
\usepackage{aaai2026} 
\usepackage{times}
\usepackage{helvet}
\usepackage{courier}
\usepackage[hyphens]{url}
\usepackage{graphicx}
\usepackage{natbib}
\usepackage{caption}
\usepackage{booktabs}
\usepackage{siunitx}
\usepackage{xspace}
\usepackage{xcolor}
\usepackage{enumitem}
\usepackage{amssymb,amsfonts,amsmath,amsthm,mathrsfs}

\nocopyright 

\title{FM4NPP: A Scaling Foundation Model for Nuclear and Particle Physics}
\author {
	David~Park\equalcontrib\textsuperscript{\rm 1},
	Shuhang~Li\equalcontrib\textsuperscript{\rm 2},
	Yi~Huang\equalcontrib\textsuperscript{\rm 1},
	Xihaier~Luo\textsuperscript{\rm 1},
	Haiwang~Yu\textsuperscript{\rm 2},
	Yeonju~Go\textsuperscript{\rm 2},
	Christopher~Pinkenburg\textsuperscript{\rm 2},
	Yuewei~Lin\textsuperscript{\rm 1},
	Shinjae~Yoo\textsuperscript{\rm 1},
	Joseph~Osborn\textsuperscript{\rm 2},
	Jin~Huang\textsuperscript{\rm 2},
	Yihui~Ren\textsuperscript{\rm 1}
}
\affiliations {
	\textsuperscript{\rm 1} AI Department, Brookhaven National Laboratory, Upton, NY\\
	\textsuperscript{\rm 2} Nuclear and Particle Physics Department, Brookhaven National Laboratory, Upton, NY\\
	\{dpark1, sli7, yhuang2, xluo, hyu, ygo, pinkenbu, ywlin, sjyoo, josborn1, jhuang, yren\}@bnl.gov
}

\newcommand{\trackml}{\texttt{TrackML}\xspace}
\newcommand{\sphenix}{\texttt{sPHENIX}\xspace}

\newcommand{\eggnet}{\texttt{EggNet}\xspace}
\newcommand{\exatrkx}{\texttt{Exa.TrkX}\xspace}
\newcommand{\fm}{\texttt{FM4NPP}\xspace}
\newcommand{\fmmvi}{\texttt{FM4NPP(m6)}\xspace}
\newcommand{\fmhead}{\texttt{AdapterOnly}\xspace}
\newcommand{\GATConv}{\texttt{GATConv}\xspace}
\newcommand{\GCNConv}{\texttt{GCNConv}\xspace}
\newcommand{\GraphConv}{\texttt{GraphConv}\xspace}
\newcommand{\SAGEConv}{\texttt{SAGEConv}\xspace}
\newcommand{\oneformer}{\texttt{OneFormer3D}\xspace}
\newcommand{\gravnet}{\texttt{GravNet}\xspace}

\newcommand{\gev}[1]{\SI{#1}{\giga\electronvolt}}

\newcommand{\pt}{\ensuremath{p_{\mathrm{T}}\xspace}}
\newcommand{\pythia} {{\textsc{pythia-8}}\xspace}
\newcommand{\GEANT}{\textsc{Geant4}\xspace}

\newcommand{\pointembeds}{spacepoint embeddings\xspace}
\newcommand{\tqueries}{track queries\xspace}
\newcommand{\tquery}{track query\xspace}
\newcommand{\rtqueries}{refined track queries\xspace}
\newcommand{\tembed}{track embedding\xspace}

\newcommand{\tinspred}{track instance prediction\xspace}
\newcommand{\td}{transformer decoder\xspace}

\newcommand{\paren}[1]{\left(#1\right)}

\begin{document}

\maketitle

\def\figurecapskip{-7pt}

\begin{abstract}

Large language models have revolutionized artificial intelligence by enabling large, generalizable models trained through self-supervision. This paradigm has inspired the development of scientific foundation models (FMs). However, applying this capability to experimental particle physics is challenging due to the sparse, spatially distributed nature of detector data, which differs dramatically from natural language. This work addresses if an FM for particle physics can scale and generalize across diverse tasks. We introduce a new dataset with more than 11 million particle collision events and a suite of downstream tasks and labeled data for evaluation. We propose a novel self-supervised training method for detector data and demonstrate its neural scalability with models that feature up to 188 million parameters. With frozen weights and task-specific adapters, this FM consistently outperforms baseline models across all downstream tasks. The performance also exhibits robust data-efficient adaptation. Further analysis reveals that the representations extracted by the FM are task-agnostic but can be specialized via a single linear mapping for different downstream tasks.

\end{abstract}

\section{Introduction}
\label{sec:introduction}

The emergence of large-scale language and vision models~\cite{Wang2023} has marked a paradigm shift from specialized neural architectures, tailored to individual tasks, toward universal, scalable, and multitasking models. These large models, containing billions of parameters and trained through self-supervised learning on massive unlabeled datasets, can be efficiently adapted to diverse downstream tasks, ranging from language translation and code generation to general reasoning. Recognizing their transformative potential, the scientific community has termed these scalable, general-purpose models as \textit{foundation models} (FMs)~\cite{bommasani2021opportunities}. Among their underpinning features, FMs can leverage self-supervised learning on extensive unlabeled datasets, allowing them to develop generalized representations adaptable to various downstream tasks with minimal additional labeled training. However, scientific data often fundamentally differ from natural language or visual data. Hence, the design and implementation of FMs for scientific fields still faces challenges~\cite{Li2024,PyzerKnapp2025}.

This work investigates developing FMs tailored for experimental nuclear and particle physics (NPP), emphasizing data from the Relativistic Heavy Ion Collider (RHIC) and the sPHENIX detector~\cite{sphenix_detector_bnl}. NPP research uses particle colliders, such as RHIC or the Large Hadron Collider (LHC), to explore subatomic phenomena. 
Discovery of the Higgs boson exemplified the transformative significance of collider-based NPP~\cite{aad2012observation}. 
In particular, RHIC collides heavy ions and polarized protons, enabling essential studies of quark-gluon plasma and the structure of protons and nuclei~\cite{Belmont:2023fau}. Commissioned in 2023~\cite{moskowitz_tiny_2023}, the sPHENIX detector features advanced tracking and calorimetry and generates extensive and complex data. The complexity of collider data and the breakthrough science it enables has motivated exploration of new data processing tools like FMs that employ self-supervised learning.

{
\setlength{\belowcaptionskip}{\figurecapskip}
\begin{figure}
	\centering
	\includegraphics[width=1\linewidth]{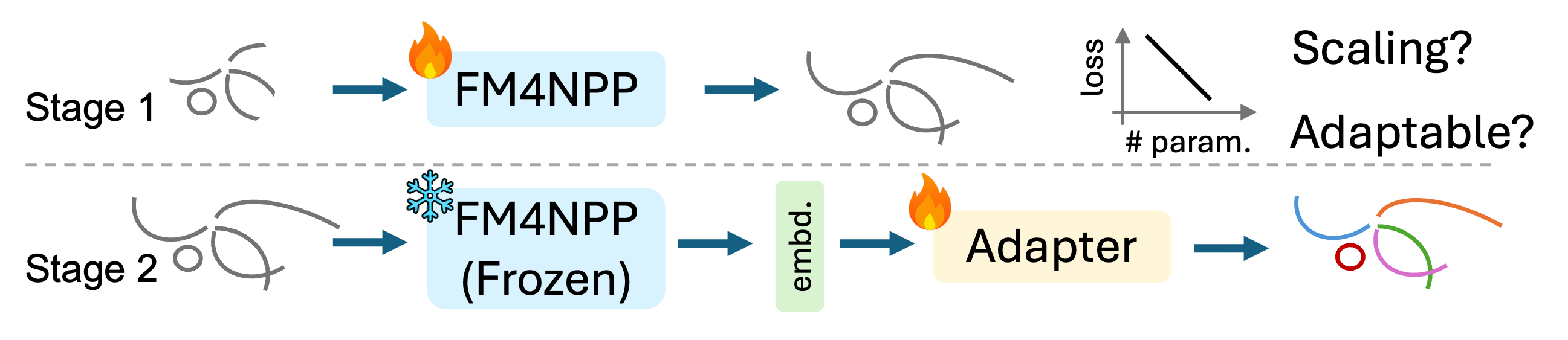}
	\caption{Overview of a scaling pretrained foundation model that can be adapted to various downstream tasks.}
	\label{fig:intro}
\end{figure}
}

However, developing an FM for NPP poses several challenges. The sparse, three-dimensional (3D)-spacepoint nature of collider data lacks an established framework for formulating self-supervised tasks.
Additionally, optimal neural architectures and the scaling behavior of pre-training losses with respect to model and data size remain unknown. Crucially, it is uncertain if neural representations from a frozen, pre-trained FM can generalize effectively to various downstream tasks, thereby outperforming existing traditional solutions and specialized AI models. 

Here, we take a first step toward enabling the use of FMs for NPP by adopting a cost-effective two-stage paradigm: (1) pretrain a large FM using a self-supervised objective and (2) pair the frozen FM with lightweight, task-specific adapters (Figure~\ref{fig:intro}). The core hypothesis is that a sufficiently trained FM encodes rich, task-agnostic representations that can be efficiently adapted to diverse downstream tasks with minimal additional training.

To this end, we construct a large-scale dataset, exceeding 11 million simulated collision events and characterized by sparse, high-dimensional detector data. We also define three downstream tasks with corresponding labeled datasets to evaluate FM adaptability. We introduce a self-supervised pre-training strategy tailored to the sparsity and structure of detector data and demonstrate strong neural scaling behavior with models up to 188 million parameters. With frozen FMs and simple adapters, we achieve state-of-the-art performance across all downstream tasks. This analysis further reveals that FM representations are broadly task-agnostic and can be specialized using a single linear transformation. In summary, our contributions are:

\begin{itemize}[itemsep=1pt]
	\item A large-scale, open benchmark dataset for FM training and evaluation in particle physics.
	\item A self-supervised pre-training method designed for sparse detector data.
	\item Empirical evidence of scaling behavior and data-efficient adaptation with frozen FMs.
	\item Insight into the structure and adaptability of FM representations across diverse tasks.
\end{itemize}

\section{Related Work}
\label{sec:related_work}

{
\setlength{\belowcaptionskip}{\figurecapskip}
\begin{figure}[t]
	\centering
	\includegraphics[width=1\linewidth]{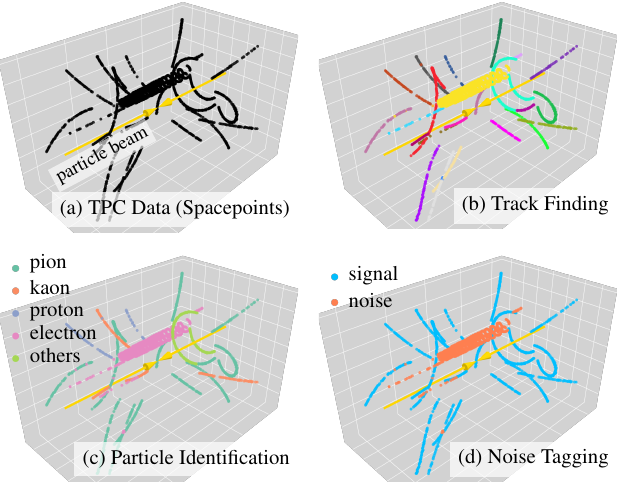}
	\caption{TPC data (spacepoints) and downstream tasks.}
	\label{fig:background}
\end{figure}
}

\paragraph{Scientific Foundation Models.} Developing FMs for scientific domains is a promising yet formidable endeavor. Progress has been most evident in domains where data exhibit modality structures similar to language or vision. For instance, Aurora~\cite{aurora} is an atmospheric FM trained on continuous spatiotemporal climate data, and recent work has demonstrated FM-based disease detection from retinal images~\cite{zhou2023foundation}. However, many scientific disciplines, including particle physics, materials science, and single-cell omics, present unique challenges like irregularly structured and sparse data. Traditional approaches, such as graph neural networks (GNNs), are well-suited for sparse data, but they face scalability issues due to phenomena, e.g., oversmoothing~\cite{rusch2023survey}. Surveys in materials science~\cite{pyzer2025foundation} and single-cell omics~\cite{ma2024harnessing} emphasize additional bottlenecks, including limited data availability and high computational costs. These challenges also apply to NPP, where it remains unclear how best to model extremely sparse data, how much data are needed, and whether pretraining benefits can effectively transfer to downstream tasks. This work takes a first step toward addressing these questions by developing a scalable FM for NPP data, focused on efficient pretraining, architectural scalability, and downstream generalization.

\paragraph{Scalable Neural Architectures.} Three neural network architectures are prominent for their scalability: Transformers, Mixture-of-Experts (MoE), and State Space Models (SSMs). The Transformer architecture~\cite{rw4} has revolutionized deep learning via self-attention, enabling effective modeling of long-range dependencies. This has led to widespread adoption in both natural language processing (NLP) and computer vision~\cite{rw5}. However, the quadratic time and space complexity of self-attention limits scalability on long sequences -- a critical bottleneck for scientific data. MoE architectures~\cite{fedus2022switch} improve inference efficiency by activating only a subset of the model per input, although they face challenges, such as training instability and expert imbalance.
The Mamba architecture~\cite{rw6}, an SSM variant, achieves linear time complexity and shows competitive or superior performance to Transformers. Given the relatively large number of spacepoints per collision event, which can result in especially long sequences, we explore SSMs as a backbone due to their favorable training efficiency and memory usage.
 
\paragraph{AI Models in NPP.} In collider physics, high-energy particles collide to produce new particles, whose trajectories—called tracks—are reconstructed from discrete spacepoints recorded by layered detector components. Track finding, or assigning spacepoints to different tracks, is one of the most important tasks. Traditional algorithms rely on combinatorial seeding followed by Kalman-filter-based refinement~\cite{rw:kf}. These classical methods are computationally expensive and difficult to parallelize on modern accelerators.
GNN‑based approaches have become popular in track finding. \exatrkx~\cite{rw:Exatrack1} formulates the task as edge classification, whereas \eggnet~\cite{rw:EggNet} employs contrastive learning followed by clustering. Each predicted track then corresponds to a connected subgraph of spacepoints.
Other recent work has introduced Transformer-~\cite{rw:trackingmaskformer} and SSM-based~\cite{rw:trackingssm} tracking models. 

While these models are promising -- with some achieving strong results using fewer than one million parameters -- there is no systematic study of scaling behavior. Moreover, open datasets designed for scaling and evaluating FMs are limited. 
In this work, more than 11 million simulated collision events are generated, affording comprehensive scaling studies. 
We also develop an FM with 188 million parameters, two orders of magnitude larger than prior models, and evaluate its performance in tracking and broader NPP tasks.

\section{Particle Detector Dataset}
\label{sec:data}

{
\setlength{\belowcaptionskip}{\figurecapskip}
\begin{figure}[t]
	\centering
	\includegraphics[width=\linewidth]{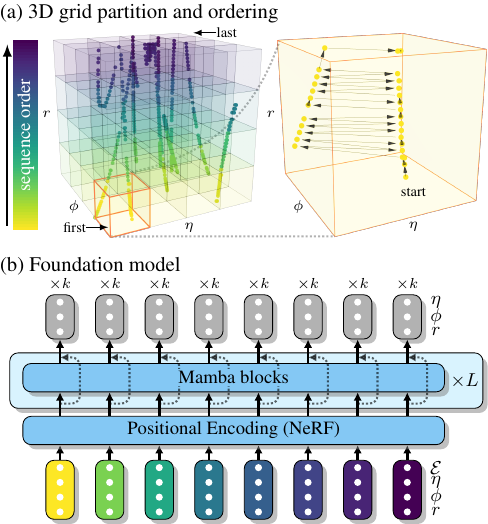}
	\caption{(a) Hierarchical Raster Scan strategy to serialize the unordered spacepoints into a 1D sequence. (b) A Mamba FM backbone for k-Next-Nearest-Neighbor prediction. }
	\label{fig:preprocess_pretrain}
\end{figure}
}

\paragraph{Dataset.}
As part of the sPHENIX detector's central tracking system, the high‑granularity Time Projection Chamber (TPC)~\cite{Klest:2020sdb} records more than $85\%$ of the total data volume. 
The TPC consists of 48 concentric cylindrical readout layers, encompassing approximately 160,000 channels that each record 260 of time samples, totaling 41.6 million voxels.
Functioning as a three-dimensional camera, the TPC records the paths of particles emerging from collision events, delivering continuous 3D spacepoint information.

We use a realistic simulation pipeline in production to generate more than 11 million proton-proton (p+p) collisions at a center-of-mass energy of $\sqrt{s}=200\,$GeV.
The natural sparsity of events in p+p collisions makes them an ideal testing ground for developing an FM for NPP applications.
The simulation pipeline includes real detector geometry, electromagnetic fields, hadronic interactions, continuous energy loss, multiple scattering, decay processes, secondary particle production, and precise energy deposition.
The raw detector hits subsequently are reconstructed to spacepoints and used as inputs in this work. 
More concretely, a collision \textbf{event} $E$ is represented as a set of \textbf{spacepoints} $\{s_i\}$, where each spacepoint is expressed by its deposit energy and location $(\mathcal{E},x,y,z)$. The number of spacepoints per event can vary from hundreds to thousands. 

\paragraph{Downstream Tasks.}  We select three complementary downstream tasks to evaluate the generalizability of an FM: \textbf{Track Finding}, \textbf{Particle Identification}~(PID), and \textbf{Noise Tagging}. \textbf{Track Finding} assigns each spacepoint to its corresponding predicted track as shown in Figure~\ref{fig:background}(a). 
Assume there are $m$ \textbf{tracks} $\{T_j\}_{j=1}^m$, where each track $T_j$ consists of its associated spacepoints $\{s_i \in T_j\}$. The goal of track finding is to predict a partition $P$ over the set of spacepoints, where $P_i^j = 1$ if spacepoint $s_i$ is assigned to track $T_j$. The number of tracks can vary from event to event. This task is analogous to instance segmentation in computer vision.

To evaluate performance, we employ both conventional physics-motivated metrics, \textbf{tracking efficiency and purity}~\cite{trackml2018}, as well as the statistical metric Adjusted Rand Index (ARI)~\cite{e:ari}. As the exact definition of whether or not a predicted track matching a true track can differ among physics experiments, we adopt the ``double-majority rule'' from the TrackML challenge~\cite{m:trackml}. 
The rule enforces that a predicted track is successfully matched to a true track only when greater than 50\% of the predicted track's spacepoints belong to that track and more than 50\% of the true track's spacepoints are present in the predicted track.
This stringent rule guarantees neither predicted tracks nor true tracks are matched at least once. 
Then, tracking efficiency (recall) is defined as the ratio between the true positive and total number of truth tracks, while tracking purity (precision) is the ratio between the true positive and total number of predicted tracks.

\textbf{PID} aims to label each spacepoint to the particle species that produced it, i.e., pion, kaon, proton, electron, and others. This is comparable to a segmentation task in computer vision. Figure~\ref{fig:background}(b) depicts an example. 
\textbf{Noise Tagging}, the third downstream task, seeks to identify spacepoints associated to low-momentum secondary particles, primarily delta electrons as they typically are not associated with physics observables of interest. This also can be considered a segmentation task.
For these two downstream tasks, we report overall accuracy, macro‐averaged precision and recall.
Additional information about the TPC detector, data generation pipeline, and statistical analysis are included in Appendix~A.

\section{Methodology}
\label{sec:method}
This section introduces the scalable FM for NPP, including a novel  serialization method for sparse spacepoints, adaptation to the Mamba architecture, and a self-supervised pretraining objective, and lightweight adapter models for downstream tasks. Additional information is included in Appendix~B.

\subsection{Self-supervised Scaling Foundation Model}
\paragraph{Serialization of Spacepoints.}
A key challenge in applying sequence-based models like Mamba to particle detector data is in serializing the unordered set of 3D spacepoints ${s_i}$ from an event $E$ into a meaningful one-dimensional (1D) sequence. 
The serialization strategy must balance two competing objectives: preserving the \textit{global} structure of particle trajectories, which typically propagate outward from the collision point, and maintaining \textit{local} continuity along individual tracks $T_j$ to retain fine-grained geometric information.
Naive serialization schemes struggle to achieve this balance. For example, space-filling curves (e.g., Hilbert or Z-order) prioritize spatial locality but often interleave points from different tracks, disrupting trajectory coherence. 
Conversely, sorting points by their radial distance preserves the outward particle flow but scatters spacepoints from the same track across distant positions in the sequence, breaking local continuity. An effective serialization must navigate this trade-off, allowing the model to learn both \textit{global} and \textit{local} physics from a sequential input.

We propose a \textit{Hierarchical Raster Scan} strategy to serialize the unordered spacepoints ${s_i}$ into a 1D sequence suitable for sequence models as shown in Figure~\ref{fig:preprocess_pretrain}(a).
First, all spacepoints are transformed from Cartesian \((x, y, z)\) to a cylindrical-polar system \((r, \phi, \eta)\) that better reflects the geometry and symmetries of collider experiments, where $r$ is radial distance, $\phi$ depicts the azimuthal angle, and $\eta$ represents the pseudorapidity (angle to the beam axis).
The raster scan method operates on two levels. The first is \textit{inter-box ordering}, where spacepoints are initially partitioned into non-overlapping 3D spatial boxes. Then, these boxes are ordered based on the $(r, \phi, \eta)$ coordinates of their geometric centers, starting from the innermost region and progressing outward. This produces a global ordering over the spatial domain. The second is \textit{intra-box ordering}, where, within each box, spacepoints are sorted by their radial coordinate $r$, which generally aligns with the direction of particle propagation.
By concatenating the intra-box sequences according to the inter-box order, we obtain a globally serialized sequence that preserves both \textit{local} spatial continuity and \textit{global} physical progression. This hierarchical structure captures important geometric and physical priors while producing a format compatible with sequence models.
Specifically, we partition the spatial domain into a $6 \times 8 \times 8$ grid along the $(r, \eta, \phi)$ axes, respectively. The $r$ bins are aligned with the physical boundaries of the TPC detector layers, while the $\eta$ and $\phi$ bins are determined using frequency-based binning to ensure balanced point distributions across the grid.

\paragraph{Mamba as a FM Model Backbone.} 
Mamba is a selective SSM that efficiently processes long sequences, achieving linear time complexity~\cite{mamba}. It features a selection mechanism that makes its internal state matrices input-dependent, allowing the model to dynamically focus on relevant information and filter out noise -- all while using a hardware-aware algorithm for fast computation. In this work, we employ Mamba2~\cite{mamba2}, which further improves upon this foundation. Mamba2 introduces structured State Space Duality (SSD), a new theoretical framework that simplifies the architecture and enhances hardware utilization, leading to significant speedups in both training and inference.

We treat every spacepoint as an input ``token'' in a sequence.
To map an input tuple $(\mathcal{E}, r, \phi, \eta)$ to the model width $d_{\text{model}}$, we employ a two-pathway process inspired by Neural Radiance Fields~\cite{nerf}. First, the feature component $\mathcal{E}$ is projected into a feature embedding of dimension $d_{\text{model}}$. Concurrently, the spatial coordinates $(r, \phi, \eta)$ are transformed with a high-frequency positional encoding function, $\gamma(\cdot)$, which uses sine and cosine transformations. Then, this encoded position is also projected into a positional embedding of dimension $d_{\text{model}}$. The final representation is the element-wise sum of the feature and positional embeddings, yielding a single vector of size $d_{\text{model}}$ that holistically captures the event's properties and location.

\paragraph{Self-supervised Pretraining Objectives.} To create a self-supervised pretraining task, the prediction objective must be decoupled from the sequence order as a naive ``next-spacepoint prediction'' would learn artifacts of the serialization itself. The target for any given spacepoint $s_i$, must be defined by its geometric relationship to other spacepoints in 3D, not its 1D sequence position. While predicting nearest neighbors is a natural geometric objective, a standard k-Nearest Neighbor task is unsuitable in an autoregressive framework due to information leakage from previously seen spacepoints. We partially address this by introducing \textit{k-Next-Nearest-Neighbor prediction} (Figure~\ref{fig:preprocess_pretrain}(b)), which aligns the objective with particle propagation. For any query spacepoint $s_i$, the model's task is to predict its $k$ spatially nearest spacepoints that reside only within its next neighborhood set $\mathcal{N}_c(s_i) = \{s_j \in E  \mid r_j > r_i \}$, which consists of spacepoints with a larger radius than $r_i$. The loss for $s_i$ is the squared Euclidean distance between the predicted coordinates of these $k$ neighbors ($\mathbf{s}_i$) and the truth coordinates ($\mathbf{y}_i$), given by $\mathcal{L}_i = ||\mathbf{s}_i - \mathbf{y}_i||_2^2$. Note that a larger $k$ means the task involves predicting more distanced spacepoints. Hence, it is inherently more difficult. 

Another obstacle in training with particle detector data is the variance in their sparsity. Rare but dense events tend to be easier to predict as points are closer to each other. This can lead to training instability. To mitigate this, we introduce an event-difficulty rescaling strategy. Events are binned by their average k-neighborhood distance (a proxy for difficulty), and the loss for each event is re-weighted by a factor corresponding to its bin. The final batch loss is the average of these re-weighted event losses, ensuring the training process is not biased by easier events.

{
\setlength{\belowcaptionskip}{\figurecapskip}
\begin{figure*}[t]
	\centering
	\includegraphics[width=.98\textwidth]{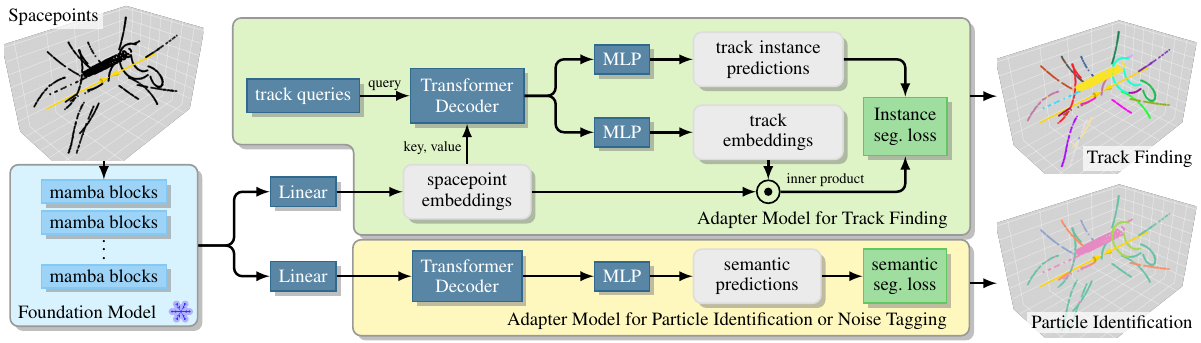}
	\caption{The pretrained FM is kept frozen during training of the adapter models for downstream tasks. The adapter models for particle identification and noise tagging share the same architecture but are trained independently.}
	\label{fig:downstream_models}
\end{figure*}
}

\subsection{Adaptive Models for Downstream Tasks}
\subsubsection{Track Finding.}
Figure~\ref{fig:downstream_models} depicts how our downstream adapter model for track finding, formulated as an instance segmentation task, is inspired by image panoptic segmentation models~\cite{rw:maskformer,rw:mask2former} and adapted to sparse spacepoints data.

Point‑level features from a FM are first projected to \pointembeds via a single linear layer, providing a probing point for the effect of the pretrained representation. We initialize $N$ learnable queries (\tqueries) $\mathbf{Q} = \{ \mathbf{q}_k \}_{n=1}^N$ and refine them over $L$ \td layers. In each layer, cross‑attention aggregates information from \pointembeds, modulated by an additive attention mask computed from intermediate assignment logits, followed by self‑attention among the queries. 
The resulting \rtqueries are passed through two separate multilayer perceptron (MLP) heads to produce a mask embedding \tembed and a classification score $\hat y_n$ . Point‑to‑query assignment probability $\hat p_{in}$ is computed as the sigmoid of the dot product between \pointembeds and each \tembed.

Let $E = \{T_j\}_{j=1}^M$ be the set of true tracks of an event $E$. We match the \rtqueries to $E$ via the Hungarian algorithm, minimizing the combined cost of Dice loss $\mathcal{L}_\text{dice}$, Focal loss $\mathcal{L}_\text{focal}$ on the per‐point assignments, and classification loss $\mathcal{L}_\text{cls}$ for track versus no‐object. 
For each matched pair $(T_j,\mathbf{q}_n)$, the loss is
$$
\mathcal{L}_\text{match}^{(j,n)}
  = \lambda_\text{dice}\,\mathcal{L}_\text{dice}^{(j,n)}
  + \lambda_\text{focal}\,\mathcal{L}_\text{focal}^{(j,n)}
  + \lambda_\text{cls}\,\mathcal{L}_\text{cls}^{(n)}.
$$
Unmatched track queries incur only $\mathcal{L}_\text{cls}^{(n)}$. We also apply auxiliary losses at each decoder layer.
At inference time, each spacepoint $i$ is assigned to the track
$
n_i^* = \arg\max_{n} \bigl(\hat p_{in} \cdot \hat y_n\bigr)
$
 and labeled accordingly.

\subsubsection{Particle Identification and Noise Tagging.}
For both PID and noise tagging tasks, illustrated in Figure~\ref{fig:downstream_models}, our lightweight adapter first projects each $d$‑dimensional point feature into a $d_p$‑dimensional embedding via a linear layer then aggregates global context with a single self-attention layer. Finally, it feeds the result through an MLP classifier.

\section{Experiments and Results}
\label{sec:experiment}
Here, we begin by examining the scaling behavior of our FM with respect to model size, dataset size, and computational cost. We then benchmark the FM paired with lightweight adapters against strong baselines across three downstream tasks. Finally, we present additional analyses to better understand the adaptation behavior of the FM.

\subsection{Neural Scaling Behaviors of FM4NPP}

\vspace{\smallskipamount}
\begin{table}[h]
	\centering
	\caption{Model Sizes and Compute Resources}\label{tab:modelsize}
	\setlength{\tabcolsep}{4pt} 
	\begin{tabular}{@{}lcccccc@{}}
		\toprule
		& \multicolumn{6}{c}{Model Sizes}\\
		\cmidrule(lr){2-7}
	 	& \texttt{m1} & \texttt{m2} & \texttt{m3} & \texttt{m4} & \texttt{m5} & \texttt{m6} \\
		\cmidrule(lr){2-7}
		Model Width & 64 & 128 & 256 & 512 & 1024 & 1536 \\
		Model Params & 0.34M & 1.3M & 5.3M & 21M & 84M & 188M \\
		\midrule
		& \multicolumn{6}{c}{Compute Resources}\\
		\cmidrule(lr){2-7}
		NVIDIA GPU & \multicolumn{2}{c}{H100 80GB} & \multicolumn{4}{c}{A100 80GB} \\
		\cmidrule(lr){2-3} \cmidrule(lr){4-7}
		Num GPUs & 1 & 1 & 4 & 8 & 24 & 64 \\
		Train Hrs & 10 & 12 & 20 & 32 & 50 & 72 \\
		\bottomrule
	\end{tabular}
\end{table}
\vspace{\smallskipamount}

{
\setlength{\belowcaptionskip}{\figurecapskip}
\begin{figure*}[t]
	\centering
	\includegraphics[width=1\textwidth]{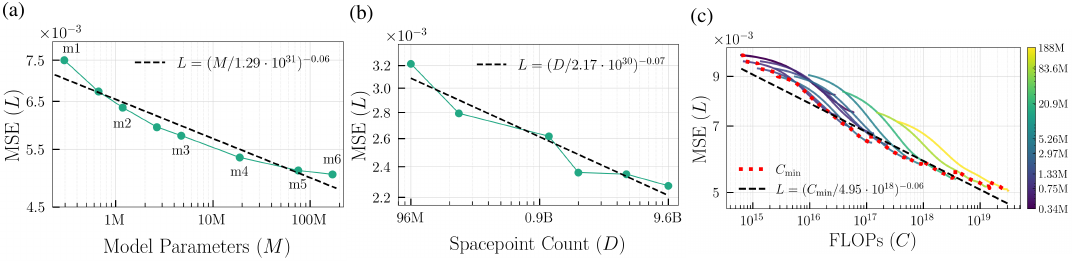}
	\caption{Neural Scaling Behaviors of FM4NPP. We evaluate neural scaling trends on increasing (a) model parameter size $M$, (b) training spacepoint count $D$, and (c) compute in FLOPs. $C_{\text{min}}$ denotes the minimum $L$ for each compute.}
	\label{fig:scaling}
\end{figure*}
}

\vspace{\smallskipamount}
\begin{table*}[ht]
	\centering
	\caption{Performance on Track Finding, Particle Identification, and Noise Tagging}
	\resizebox{\linewidth}{!}{
		\begin{tabular}{l r ccc | l r ccc ccc}
			\toprule
			&  & \multicolumn{3}{c|}{Track Finding} & & & \multicolumn{3}{c}{Particle Identification} & \multicolumn{3}{c}{Noise Tagging} \\
			\cmidrule(lr){3-5} \cmidrule(lr){8-10} \cmidrule(lr){11-13}
			model & \#trnbl para. & ARI$\uparrow$ & efficiency$\uparrow$ & purity$\uparrow$ & model & \#trnbl para. & acc.$\uparrow$ & recall$\uparrow$ & pre.$\uparrow$ & acc.$\uparrow$ & recall$\uparrow$ & pre.$\uparrow$\\
			\midrule
			\eggnet  & $0.16$M & $0.7256$ & $74.19\%$ & \underline{$75.14\%$} & \SAGEConv & $0.91$M & $0.7262$ & $0.4563$ & \underline{$0.6502$} & $0.9174$ & $0.7227$ & $0.8165$ \\
      \exatrkx & $3.86$M & \underline{$0.8765$} & \underline{$91.79\%$} & $66.42\%$ & \oneformer & $44.95$M & \underline{$0.7701$} & \underline{$0.4897$} & $0.5767$ & \underline{$0.9646$} & $\mathbf{0.9404}$ & \underline{$0.8948$} \\
			\fmhead  & $2.39$M & $0.7243$ & $78.01\%$ & $64.54\%$ & \fmhead & $0.74$M & $0.6631$ & $0.3387$ & $0.6111$ & $0.9111$ & $0.6215$ & $0.8359$ \\
			\fmmvi     & $2.39$M & $\mathbf{0.9448}$ & $\mathbf{96.08\%}$ & $\mathbf{93.08\%}$ & \fmmvi  & $0.74$M     & $\mathbf{0.9039}$ & $\mathbf{0.7652}$ & $\mathbf{0.8782}$ & $\mathbf{0.9713}$ & \underline{$0.9367$} & $\mathbf{0.9190}$ \\
			\bottomrule
		\end{tabular}
	}
	\label{tab:performance}
\end{table*}
\vspace{\smallskipamount}

We evaluate our FM's scaling behavior across three axes: model size, dataset size, and compute budget. Results are summarized in Figures~\ref{fig:scaling}(a–c).
\paragraph{Model Scaling.}
We construct a series of FMs with varying capacities, denoted \texttt{m1} through \texttt{m6} in Table~\ref{tab:modelsize}. 
Figure~\ref{fig:scaling}(a) shows the validation mean squared error~(MSE) plotted against model size on a log-log scale, revealing a clear power-law relationship. 
As the number of parameters increases, validation loss consistently decreases, which aligns with neural scaling laws observed in language and other scientific domains~\cite{nlpscale2,nlpscale1,climax,aurora}. Notably, performance plateaus at \texttt{m6}, suggesting a possible saturation point, which we leave for future investigation.
\subsubsection{Data Scaling.}
To isolate the effect of training dataset size, we train the \texttt{m3} model on varying subsets (1\%, 2.4\%, 11.6\%, 20\%, 47.6\%, 100\%) of the full dataset. Figure~\ref{fig:scaling}(b) shows how performance improves steadily with more data, again following a power-law trend. This suggests the FM can continue to benefit from the large-scale data routinely produced in collider experiments.
\paragraph{Compute Scaling.}
Finally, we study the relationship between compute and model performance. Validation MSE is plotted against the total number of floating-point operations (FLOPs) used during training (Figure~\ref{fig:scaling}c). Models up to \texttt{m3} are trained at 25\%, 50\%, and 100\% of their total iteration budget, while larger models (\texttt{m4}-\texttt{m6}) are trained to full completion only. The results show that smaller models are initially more compute-efficient, but larger models outperform them when more resources are allocated. This highlights the importance of compute-optimal model scaling for deployment in high-throughput environments.
All experiments have been conducted using A100 and H100 GPUs with corresponding hardware costs summarized in Table~\ref{tab:modelsize}.

All models are trained with a batch size of 256. An optimal learning rate of $2 \times 10^{-4}$ is selected through hyperparameter tuning on the \texttt{m3} model and reused across all variants using the \textmu-parameterization principle~\cite{mumamba}. This approach ensures consistent gradient flow across model sizes and enables zero-shot hyperparameter transfer~\cite{mufive}. Smaller models (\texttt{m1}, \texttt{m2}) are trained for 50,000 iterations, while larger models (\texttt{m3}–\texttt{m6}) are trained for 100,000 iterations. We apply cosine learning rate decay with 10,000-step linear warmup and use gradient clipping at a threshold of 0.1. All experiments employ the AdamW optimizer~\cite{adamw} with a weight decay of 0.01. Additional training details are provided in Appendix~B.

\subsection{Performance on Downstream Tasks}
    
\paragraph{Track Finding.} 
For baseline models used for track finding, we select \exatrkx~\cite{rw:Exatrack1} and \eggnet~\cite{rw:EggNet}. Both models are GNN-based and designed specifically for track finding. However, due to the difference in detector geometry, we must adapt these methods for our data (details are documented in Appendix~C).
To verify embeddings extracted using the pretrained FM provide richer information, we also train the lightweight adapter model alone. 

Table~\ref{tab:performance} reports the track finding results of our FM with several baselines. All metrics are computed over the entire test set rather than averaged per event. For example, tracking efficiency is defined as the fraction of all true tracks in the dataset that are successfully matched. 
Our model achieves higher performance on conventional clustering metrics such as ARI, and also outperforms other approaches in tracking efficiency (recall) and purity (precision).

We also compare this work against the official sPHENIX reconstruction pipeline, which employs a Cellular Automaton seeding followed by a Kalman filter~\cite{Osborn:2021zlr}. As that algorithm is optimized for high transverse momentum (\pt), long tracks within the TPC acceptance, we restrict this comparison to tracks that leave at least 20 spacepoints in the TPC and satisfy $\pt>\gev{1}$ and $|\eta|<1.1$. Under these criteria, our model reaches a tracking efficiency of 99.6\%, exceeding the sPHENIX pipeline's 94.6\%.

\paragraph{Particle Identification and Noise Tagging.}

For the PID and noise tagging tasks, we experiment with four conventional GNN models and report on the best performing one, \SAGEConv. The graph edge set is constructed by $k$-nearest neighbors with a distance cap. We also adapt and train a state-of-the-art segmentation model for 3D point cloud data named \oneformer~\cite{kolodiazhnyi2024oneformer3d}. 

Table~\ref{tab:performance} reports the segmentation accuracy, as well as macro‐averaged recall and precision. For the PID task, our FM consistently outperforms all baselines, achieving the highest accuracy, recall, and precision. Meanwhile, for the noise tagging, the FM outperforms all GNN-based baselines with similar performance compared to \oneformer. It worth noting that \oneformer has about $45$ million trainable parameters, whereas our adapter head has $0.74$ million. More details about comparative model implementations and sample outputs are provided in Appendix~C.

\subsection{Insights about FM Adaptation}
We aim to further understand FM adaptation behaviors by answering the following research questions:
\begin{itemize}[itemsep=1pt]
    \item \textbf{Q1}: Does increasing the size of the FM lead to improved performance on downstream tasks?
    \item \textbf{Q2}: Do larger FMs require fewer labeled examples to achieve comparable performance (i.e., better data efficiency)?
    \item \textbf{Q3}: Are the learned FM embeddings task-agnostic, and, if so, how much adaptation is needed to specialize them for specific tasks?
\end{itemize}

{
\setlength{\belowcaptionskip}{\figurecapskip}
\begin{figure}[ht]
	\centering
	\includegraphics[width=0.95\linewidth]{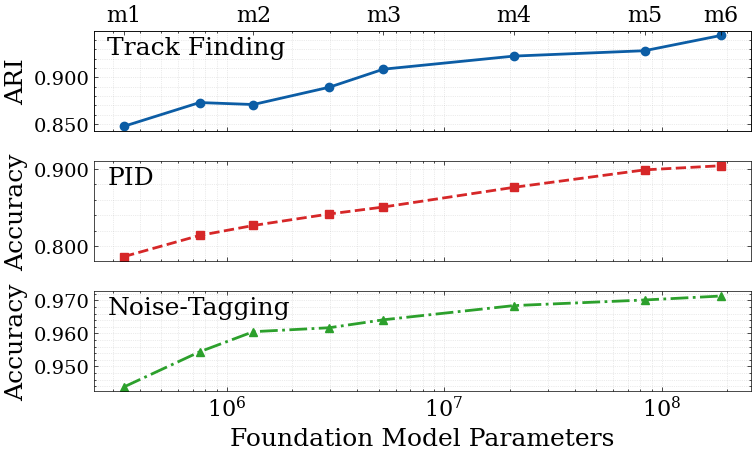}
	\caption{Effect of FM Model Size on Downstream Task Performance.}
	\label{fig:downstream_vs_model_size}
\end{figure}
}

{
\setlength{\belowcaptionskip}{\figurecapskip}
\begin{figure}[ht]
	\centering
	\includegraphics[width=0.95\linewidth]{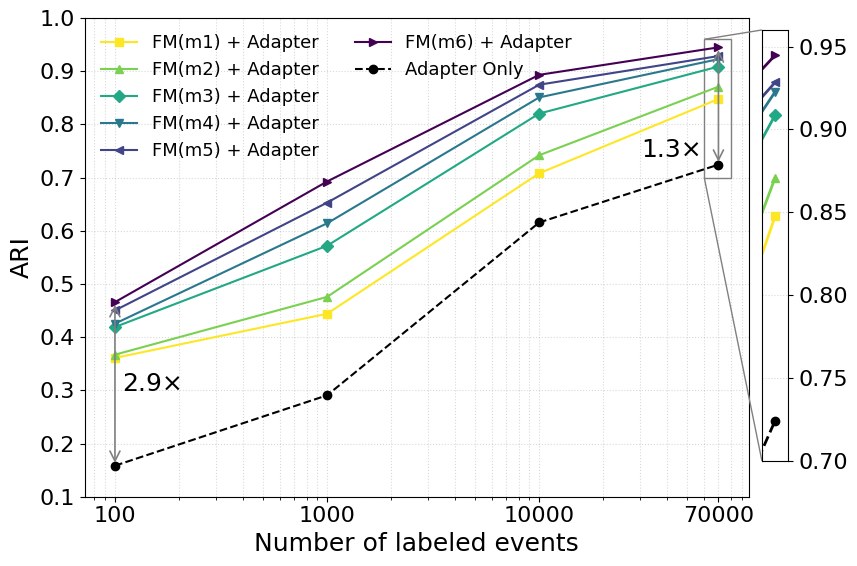}
	\caption{Larger FM is more data efficient.}
	\label{fig:downstream_vs_data_size}
\end{figure}
}

Figure~\ref{fig:downstream_vs_model_size} shows the downstream performance of all three tasks plotted against a pretrained FM size. All tasks use the same frozen pretrained representation. Larger pretrained models consistently yield higher performance across every task, confirming that scaling up the pretrained model size improves on various downstream performance even when only the lightweight decoder head is trained.

Figure~\ref{fig:downstream_vs_data_size} depicts the ARI for the track finding task training on different number of labeled data, from 100 to 70,000. Larger FMs consistently outperform smaller ones across different levels of labeled data, indicating neural embeddings extracted from larger FMs contain richer information and can be generalized easier. This confirms common empirical observations that larger models can generalize better~\cite{novak2018sensitivity}. In addition, compared to a baseline adapter-only model trained solely on labeled data (dashed line in Figure~\ref{fig:downstream_vs_data_size}), pretraining a self-supervised FM model on a large amount of unlabeled data -- easy to come by in NPP -- proves quite effective. The relative gain in ARI is greater in the fewer labeled data regions compared to those with abundant labeled data: $2.9\times$ versus $1.3\times$. 

{
\setlength{\belowcaptionskip}{\figurecapskip}
\begin{figure}[ht]
	\centering
	\includegraphics[width=\linewidth]{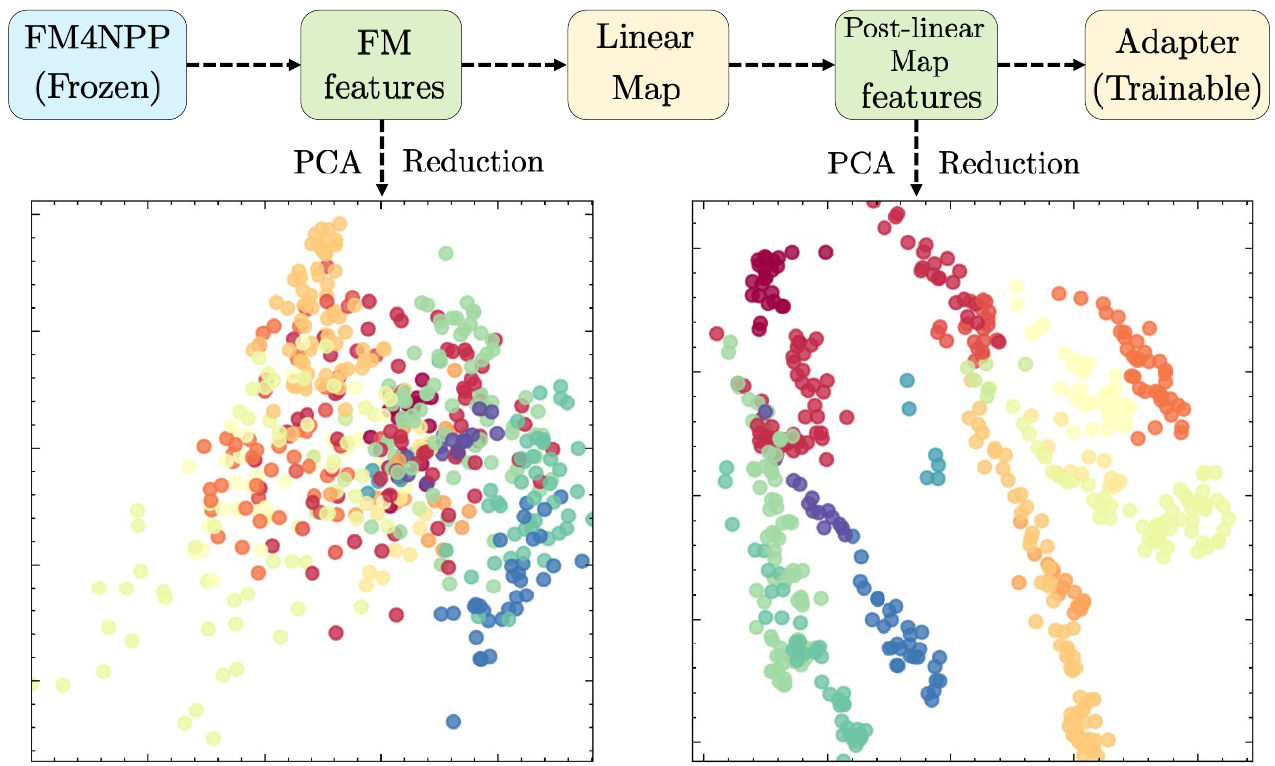}
	\caption{Visualization of learned embeddings from the FM (left) and the post-linear map features (right), projected via PCA reduction. Each marker corresponds to a spacepoint, colored by its associated track identity.}
	\label{fig:XAI}
\end{figure}
}

We analyze the neural embeddings from the frozen FM and their transformation after a simple linear projection, which precedes the lightweight adapter used for downstream tasks. 
To assess the task specificity of these features, we apply dimensionality reduction techniques (e.g., PCA) to both the raw FM embeddings and linearly projected features, focusing on a representative downstream task: track finding.
As shown in Figure~\ref{fig:XAI}, the raw FM embeddings exhibit no clear separation among particle tracks, indicating the representations are task-agnostic. However, after applying a single linear projection, distinct and well-separated clusters emerge, corresponding to different particle tracks.
This finding indicates the FM encodes rich, general-purpose representations that require minimal adaptation for task-specific use. It also explains why lightweight adapters, when built atop FM embeddings, outperform non-FM baselines by leveraging semantically meaningful input features. 
Additional analyses are provided in Appendix~C.

\section{Conclusion and Future Work}
\label{sec:discussion}

With this work, we demonstrate that FMs can be effectively extended to experimental particle physics by introducing a scalable self-supervised training strategy tailored to sparse detector data. Our model, trained on more than 11 million events, generalizes across diverse downstream tasks with frozen weights and lightweight adapters, consistently outperforming task-specific baselines. Its effective performance and data efficiency suggest the model learns rich, task-agnostic representations that are easily adapted using simple mappings. These findings reveal the potential for general-purpose, scalable models in NPP.

Future work will explore scaling behavior in greater depth by investigating larger models, larger datasets, and increased computational budgets, as well as  alternative architectures. It may also be valuable to extend the FM paradigm to incorporate supervised fine-tuning jointly across multiple downstream tasks. 
On the application side, incorporating additional detector subsystems—such as calorimeters and silicon trackers—could enable broader downstream tasks, including particle-flow reconstruction and rare event tagging~\cite{atlas_bjet}.
Another open question is whether the FM paradigm can generalize to heavy-ion collisions or be unified across multiple collider experiments worldwide.

\clearpage 

\section{Acknowledgments}
The authors would like to express their sincere gratitude to the sPHENIX Collaboration for sharing the simulation data and experimental knowledge, as well as Jubin Choi, Abhay Deshpande, Alexei Klimentov, Michael Begel, Torre Wenaus, Nicholas D'Imperio, James Dunlop, and John Hill from Brookhaven National Lab for their valuable support and feedback.

This work was supported by the Laboratory Directed Research and Development (LDRD) Program at Brookhaven National Laboratory, LDRD 25-045, which is operated and managed for the U.S. Department of Energy (DOE) Office of Science by Brookhaven Science Associates under contract No. DE-SC0012704.
Shuhang Li was partially supported by the DOE Office of Science through the Office of Nuclear Physics under Award No.~DE‐FG02‐86ER40281.
Yihui Ren, Xihaier Luo and Shinjae Yoo were partially supported by  the DOE Office of Science through the Office of Advanced Scientific Computing Research and the Scientific Discovery through Advanced Computing (SciDAC) program.

This research also utilized resources of the National Energy Research Scientific Computing Center (NERSC), a DOE Office of Science User Facility, under NERSC Award No. DDRERCAP-m4722. The authors are grateful to the NERSC staff for their support, particularly Shashank Subramanian and Wahid Bhimji.

\bibliography{bib}

\clearpage

\onecolumn
\begin{center}
  {\LARGE \bfseries Appendix for \\[0.5ex]
    “FM4NPP: A Scaling Foundation Model for Nuclear and Particle Physics”}
\end{center}
\vspace{1em}

\appendix 

\section{Dataset}
\label{app:dataset}
The dataset used in this work is based on simulated proton–proton (p+p) collisions at center‑of‑mass energy of $\sqrt{s}=200\,$GeV, corresponding to the conditions of the sPHENIX experiment at the Relativistic Heavy Ion Collider (RHIC), where charged-particle trajectories are recorded with the Time Projection Chamber (TPC). p+p collisions serve as a precision workhorse for testing QCD and nucleon structure and provide the baseline for quantifying how particle production in heavy‑ion collisions, viewed as a superposition of pp interactions, is modified by the QGP~\cite{m:HIintro}.

{
\setlength{\belowcaptionskip}{\figurecapskip}
\begin{figure}[ht]
	\centering
    	\includegraphics[width=\linewidth]{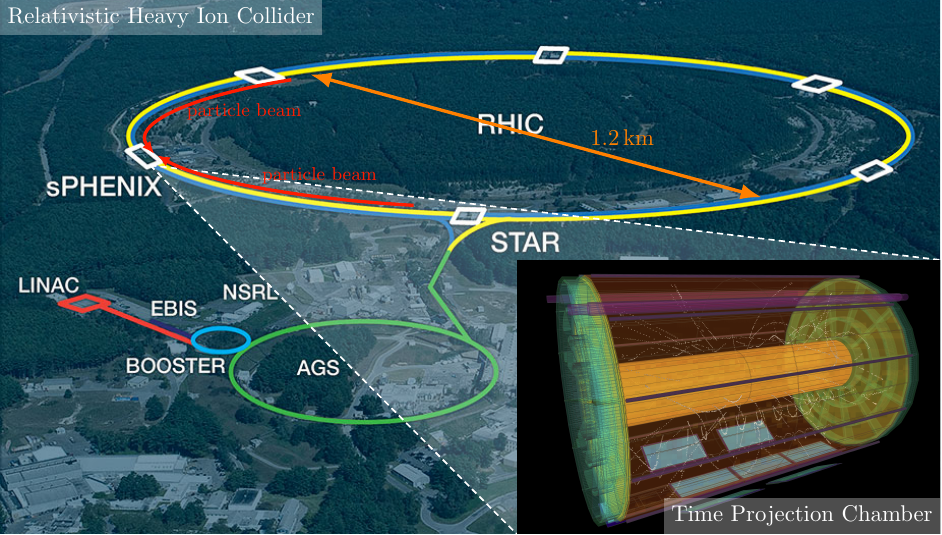}
    	\caption{Relativistic Heavy Ion Collider at Brookhaven National Lab and sPHENIX Experiment.}
    	\label{fig:app-rhic}
\end{figure}
}

\subsection{Simulation and Provenance}
Minimum‑bias p+p collisions are generated with \pythia.307~\cite{m:pythia8} `Detroit' tune~\cite{PYTHIA_tune}, and then propagated through a full {\GEANT}~\cite{m:g4} simulation of the as‑built sPHENIX detector, including its detailed CAD geometry and measured $1.4$T field map. We employ the `FTFP\_BERT\_HP' physics list for high‑precision treatment of neutron and hadron interactions. The simulation chain models continuous energy loss, multiple scattering, secondary particle production, and decay processes with the true material budget, supports space‑charge distortion and its data‑driven correction, and carries signals through the full front‑end electronics (shaping, digitization, zero suppression, and channel‑by‑channel gain/noise).

The simulated TPC response, so‑called \textsc{G4Hits}, emulates raw ionization signals from charged particles traversing the TPC volume, which are reconstructed into spacepoints reflecting the true spatial resolution and distortions. Each spacepoint is then matched to the Monte Carlo truth particle that produced it, and the particle’s properties—identity, momentum, and track association—are recorded as ground‑truth labels for our downstream tasks (track finding, PID, and noise identification).

All of the code used to drive the \pythia{} and {\GEANT} simulations, as well as the downstream emulation and reconstruction chain (digitization, spacepoint reconstruction), is sourced from the sPHENIX software stack, including the core simulation and reconstruction libraries and supporting infrastructure~\citep{sphenix_coresoftware,sphenix_acts,sphenix_macros,sphenix_calibrations}. 

\subsection{Contents and Structure}

Each event contains:
\begin{itemize}
  \item Reconstructed spacepoints from the TPC, including position and ionization energy.
  \item Monte Carlo truth particles with their PDG identity, momentum at production, and vertex location at production.
  \item Associations between spacepoints and truth particles.
\end{itemize}

\subsection{Dataset Statistics}
The event-level complexity in the dataset varies widely. As shown in Fig~\ref{fig:app-dist}, the number of reconstructed TPC spacepoints per event ranges from a few hundred to tens of thousands, reflecting low-multiplicity to relatively busy collision topologies. Correspondingly, the number of truth tracks per event spans from under ten up to nearly one hundred.

\begin{figure}[ht]
    \centering
    \includegraphics[width=.5\linewidth]{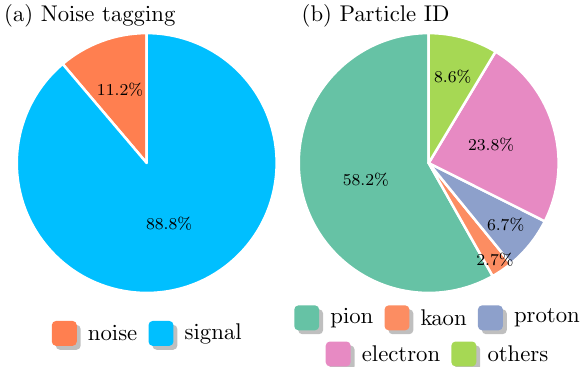}
    \caption{Class ratio of noise tagging and particle identification.}
    \label{fig:app-class_ratio}
\end{figure}

Figure~\ref{fig:app-class_ratio} summarizes the class composition for the noise-tagging and particle identification (PID) downstream tasks.

{
\setlength{\belowcaptionskip}{\figurecapskip}
\begin{figure}[h]
    \centering
    \includegraphics[width=\linewidth]{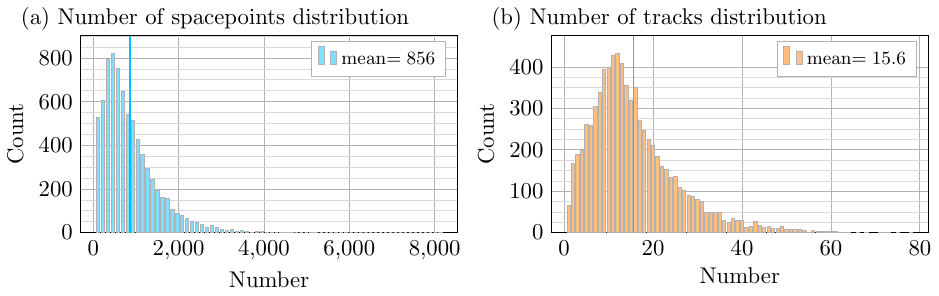}
    \caption{Distributions of number of spacepoints and tracks per event.}
    \label{fig:app-dist}
\end{figure}
}

\paragraph{Noise-tagging.} 
Noise spacepoints are defined operationally based on the truth-level kinematics of their progenitor particles. 
Specifically, any spacepoint associated with a Monte Carlo truth track whose momentum is below \(60\ \mathrm{MeV}/c\) is labeled as noise. 
Particles produced in the primary p+p collision with such low momentum are kinematically unable to reach the active TPC volume due to the magnetic field; therefore, spacepoints matched to these low-momentum tracks arise predominantly from secondary interactions with detector material (e.g., delta electrons, conversion products, or other material-induced processes). 
These secondary-origin spacepoints are not part of the primary signal topology of interest and are treated as “noise” for the purposes of the corresponding downstream classification task. 

\paragraph{PID.} 

The PID task uses five coarse-grained target classes, grouping charge-conjugate species together to reduce sparsity while preserving physics relevance:
\begin{itemize}
  \item \textbf{Pion:} \(\pi^+\), \(\pi^-\)
  \item \textbf{Kaon:} \(K^+\), \(K^-\)
  \item \textbf{Proton:} proton and anti-proton
  \item \textbf{Electron:} electron and positron
  \item \textbf{Other:} all remaining particle species 
\end{itemize}

The class ratios shown in Fig.~\ref{fig:app-class_ratio} reflect the inherent imbalance in these labels, driven by the underlying particle production spectra and the noise definition.

\section{Methodology}
\label{app:methodology}

\subsection{Preliminaries} 
This section outlines a compact mathematical way to express the hierarchical relationship between events, tracks, and spacepoints in a particle detector like a Time Projection Chamber (TPC). A collision \textbf{event} $E$ is represented as a set of \textbf{tracks} $\{T_j\}$, where each track $T_j$ is an ordered sequence of \textbf{spacepoints} $(s_k)$, and each spacepoint $s$ is a vector $(E_{\text{dep}}, x, y, z, \dots)$ containing its physical properties. Concretely, we express a single event, $E$, as follows:
$$ E = \{ T_j \}_{j=1}^{m} $$
This states that an event ($E$) is a set containing $m$ individual tracks ($T_j$). The number of tracks, $m$, is variable for each event. Each track, in turn, is defined by its constituent spacepoints:
$$ T_j = (s_{j,k})_{k=1}^{n_j} $$
This expresses that a single \textbf{track ($T_j$)} is an ordered sequence of $n_j$ spacepoints ($s_{j,k}$). The sequence is ordered because particles follow a specific path through the detector, and the number of spacepoints per track, $n_j$, is also variable. Finally, each individual spacepoint is a vector of its properties, which can be represented abstractly as:
$$ s_{j,k} \in \mathbb{R}^D $$
A \textbf{spacepoint ($s$)} is a vector in a D-dimensional feature space. \noindent A \textbf{Spacepoint ($s_{j,k}$)} is now explicitly defined as a vector containing its primary physical properties:
$$ s_{j,k} = (\mathcal{E}, x, y, z)_{j,k} $$ where $\mathcal{E}$ is the energy deposited by the particle at that point in the detector, and \textbf{$(x, y, z)$} is the spatial coordinates of the spacepoint.

\subsection{Coordinate Transformation}
 We transform spacepoint coordinates from Cartesian \((x, y, z)\) to a cylindrical-polar system \((r, \phi, \eta)\) that better reflects the geometry and symmetries of collider experiments. The radial distance $r$ is defined as \(r = \sqrt{x^2 + y^2}\), measuring how far a point lies from the beamline in the transverse plane, and is essential for evaluating transverse momentum and energy. The azimuthal angle $\phi$ is given by \(\phi = \operatorname{atan2}(y, x)\), describing the orientation of the spacepoint in the \(x\)-\(y\) plane and exploiting the detector’s cylindrical symmetry around the beam axis. The pseudorapidity $\eta$ is defined as \(\eta = -\ln\left[\tan\left(\theta/2\right)\right]\), where \(\theta = \operatorname{atan2}(r, z)\) is the polar angle; this coordinate is used instead of \(\theta\) because particle production tends to be uniform in \(\eta\), and for highly relativistic particles, \(\eta\) approximates the Lorentz-invariant rapidity. Finally, to ensure consistent feature scaling, we apply a min-max normalization to the spatial coordinates, transforming the pseudorapidity ($\eta \in [-2, 2]$), azimuthal angle ($\phi \in [-\pi, \pi]$), and radial distance ($r \in [30, 78]$, centimeters) into the interval $[0, 1]$. The transformed $s_i = (\mathcal{E}, r, \phi, \eta)_i$ are used for all analyses described in this paper.

\subsection{Serialization} 
Our objective is to perform self-supervised pretraining on the raw 3D point cloud of particle spacepoints from a collision event, $S = \{s_1, \dots, s_N\}$. To leverage the power of sequential models like MAMBA, which have excelled in learning rich representations, we must first solve the fundamental problem of transforming the unordered 3D set into an ordered 1D sequence. This \textbf{serialization} process is not merely a technical step; the choice of ordering scheme is critical to preserving the underlying physical structure of the data.

An ideal serialization must satisfy two competing demands: it must respect the \textit{global} physics of the event (i.e., particles flying outwards) while simultaneously preserving the \textit{local} continuity of individual particle tracks.

We first analyze and dismiss naive approaches. A space-filling curve, for example, excels at preserving 3D locality but completely disregards the concept of a track; its path erratically jumps between physically distinct trajectories, creating a chaotic signal. Conversely, a simple global raster scan on the spacepoints' cylindrical coordinates, $s'_i = (r_i, \phi_i, \eta_i)$, respects the outward propagation along the radius but fails on local continuity. The initial hits of a track (at low $r$) become ``context-starved,'' as their preceding elements in the sequence belong to entirely different tracks.

\paragraph{\underline{Proposed Solution: Hierarchical Raster Scan}}

To resolve this dichotomy, we introduce a \textbf{Hierarchical Raster Scan}. This method balances global structure with local context by operating on two levels:
\begin{enumerate}
    \item \textbf{Partitioning:} The entire detector volume is partitioned into a grid of smaller 3D ``boxes.''
    \item \textbf{Ordering:} A raster scan using the physically-motivated order $(r, \phi, \eta)$ is applied twice. First, it orders the spacepoints \textit{within} each box (intra-box ordering). Second, it orders the boxes themselves based on their geometric centers (inter-box ordering).
\end{enumerate}

This strategy ensures that the sequence progresses globally outwards but maintains local contiguity within each partitioned region. However, even with this optimal serialization, a profound challenge remains. If the learning objective were to simply predict the next hit in this sequence, the model would be forced to learn the arbitrary artifacts of the serialization itself, particularly the artificial jumps at box boundaries.

Therefore, designing a robust serialization scheme is a necessary but insufficient step. The learning objective must be intelligently designed to be independent of these serialization artifacts, a challenge we address in the subsequent section.

\paragraph{\underline{Physics Informed Partitioning} }
The division of the detector volume into a grid is not uniform; it is a physics-informed partitioning designed to align with both the detector's physical geometry and the observed distribution of particle hits. This ensures the partitioning itself provides a meaningful structural prior for the learning task.

For the azimuthal angle ($\phi$) and pseudorapidity ($\eta$) dimensions, the binning is data-driven. The boundaries are specifically chosen to create bins with a roughly uniform density of hits. This strategy balances the information content across partitions, preventing high-occupancy regions from disproportionately influencing the model. A detailed number of bins and illustration of this binning strategy is provided in Figure~\ref{fig:binning}.

{
\setlength{\belowcaptionskip}{\figurecapskip}
\begin{figure}[ht]
    \centering
    \includegraphics[width=\linewidth]{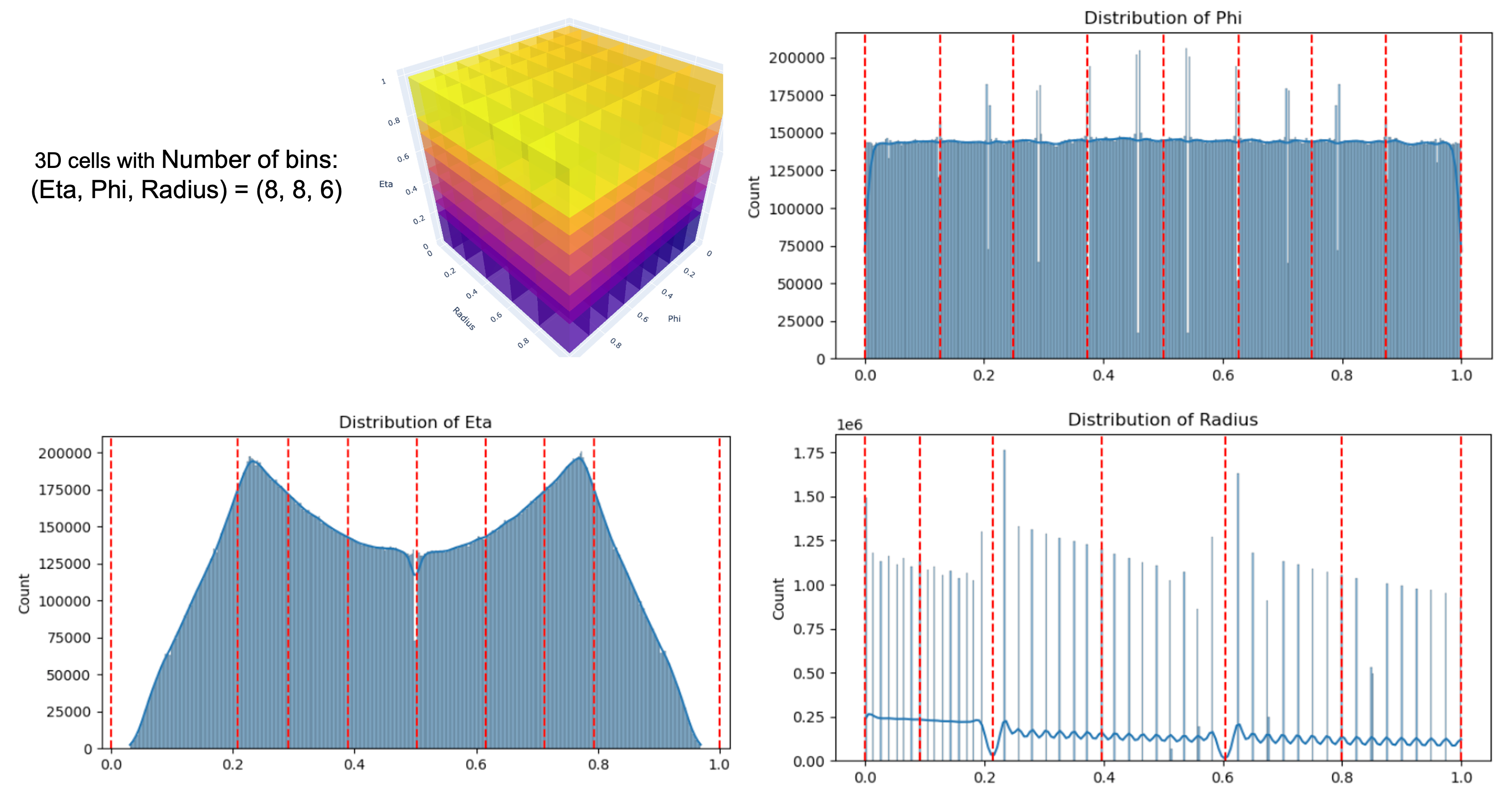}
    \caption{Physics Informed Partitioning. Top-left graph shows the binning of the data space into 384 bins ($8\times8\times6$). The other plots show the distribution of spacepoint values in normalized Phi, Eta, and Radius dimensions, respectively, computed using 50,000 events.}
    \label{fig:binning}
\end{figure}
}

For the radial dimension ($r$), the partitioning mirrors the physical construction of the sPHENIX detector. The detector's 48 layers are arranged in three major groups; therefore, we create six radial bins, allocating two bins to each major detector group. By embedding the detector's known layered structure into the partitioning scheme, we further ground the serialization process in the experiment's physical reality.

\subsection{MAMBA: Selective State Space Models}
\label{app:mamba}
 Mamba represents a significant advancement in sequence modeling, challenging the dominance of the Transformer architecture, particularly for long sequences. It is a selective state space model (SSM) that combines the strengths of recurrent neural networks (RNNs) and convolutional neural networks (CNNs) to offer linear-time complexity and constant-time inference.

Mamba's foundation is the State Space Model, a continuous-time system described by the following linear ordinary differential equation:
\begin{equation*}
\begin{aligned}
    \frac{dh(t)}{dt} &= Ah(t) + Bx(t), \\
    y(t) &= Ch(t) + Dx(t)
\end{aligned}
\end{equation*}
Here, $h(t)$ is the latent state, $x(t)$ is the input, and $y(t)$ is the output. $A$, $B$, $C$, and $D$ are matrices that are typically learned from data.

For use in deep learning, this continuous system is discretized. A crucial step in Mamba is making the key matrices, particularly the transition matrix $A$ and the input projection matrix $B$, selective and input-dependent. This is achieved by having dedicated neural networks that predict these matrices based on the current input token.

The discretized formulation of the state transition is:
\begin{equation}
    h_t = \bar{A}h_{t-1} + \bar{B}x_t
\end{equation}
Where $\bar{A}$ and $\bar{B}$ are the discretized, input-dependent matrices. This selectivity allows Mamba to modulate its recurrent state, effectively controlling how much of the past to retain and how to incorporate the current input. The model can be unrolled for efficient parallel training, similar to a CNN, or used in a recurrent manner for constant-time inference.

\paragraph{\underline{Architectural Principles}} The core innovation of Mamba lies in its selective mechanism, which allows the model to dynamically adapt its parameters based on the input. This enables it to focus on relevant information and filter out noise, a crucial capability for processing long and complex sequences. Unlike traditional SSMs, which are time-invariant, Mamba's parameters are functions of the input, making it a time-varying system. Key components of the Mamba architecture include:
\begin{itemize}
    \item Selective State Space Layer: This is the fundamental building block of Mamba. It replaces the attention mechanism and feed-forward network of a Transformer block.\\    
    \item Hardware-Aware Algorithm: Mamba employs a parallel scan algorithm that is optimized for modern hardware (GPUs), enabling efficient training and inference. This algorithm avoids the materialization of the full state sequence, a significant memory bottleneck in traditional SSMs.
\end{itemize}

\paragraph{\underline{Mamba2}} Mamba2 is a direct successor to Mamba, designed to further improve upon its efficiency and performance. It introduces a new theoretical framework called Structured State Space Duality (SSD), which provides a deeper understanding of the relationship between SSMs and other architectures like Transformers. The primary motivation behind Mamba-2 was to address some of the hardware utilization inefficiencies of the original Mamba. While Mamba offered linear-time complexity, its performance on modern GPUs could still be optimized. Key improvements in Mamba-2 include:
\begin{itemize}
    \item State Space Duality (SSD): This framework establishes a formal equivalence between a class of structured SSMs and a form of global convolution. This duality allows for the design of more efficient algorithms by leveraging insights from both perspectives.\\
    \item Architectural Simplifications: Mamba-2 simplifies the Mamba block by replacing the complex selective scan with a more structured and hardware-friendly formulation derived from the SSD framework. This often involves a multi-headed Mamba block, analogous to the multi-head attention in Transformers.\\
    \item Improved Hardware Utilization: The redesigned architecture of Mamba-2 is more amenable to parallelization on modern hardware, leading to significant speedups in both training and inference compared to the original Mamba.
\end{itemize}

\subsection{FM4NPP: Architecture}
\paragraph{\underline{Positional Embedding}}
The model first transforms the raw input data into a high-dimensional space suitable for sequence processing. An input batch of serialized collision events is represented as a tensor of shape $(B, S, 4)$, where B is the batch size, S is the sequence length, and each spacepoint is a 4-dimensional vector comprising its deposited energy and 3D spatial coordinates $(E_{\text{dep}}, \eta, \phi, r)$. This tensor is processed by an embedding module that projects the 4D spacepoint features into the model's latent space, $D_{\text{model}}$. It also computes a positional encoding from the 3D spatial coordinates using a function $\gamma(\cdot)$ inspired by Neural Radiance Fields (NeRF), defined as:
\begin{equation*}
    \gamma(\mathbf{p}) = \left( \mathbf{p}, \sin(2 \mathbf{p}), \cos(2 \mathbf{p}), \dots, \sin(2^l \mathbf{p}), \cos(2^l \mathbf{p}) \right)
\end{equation*}
where $\mathbf{p}$ is the coordinate vector and the frequencies $2^l$ are sampled from a geometric progression. This encoding, also mapped to $D_{\text{model}}$, is combined with the feature representation via element-wise addition. The output of this stage is a single tensor of shape $(B, S, D_{\text{model}})$, where $D_{\text{model}}$ is the model width.

\paragraph{\underline{Network Architecture and k-Next Nearest Neighbor Prediction Head}}
The core architecture consists of a stack of Mamba blocks that sequentially process the embedded hits. The input to the first block is the $(B, S, D_{\text{model}})$ tensor from the embedding stage. Each block operates as follows:
\begin{itemize}
    \item Pre-Normalization: The input tensor is first passed through a Root Mean Square Normalization (RMSNorm) layer. This layer normalizes the feature vector of each spacepoint independently.\\
    
    \item Sequence Modeling: The normalized $(B, S, D_{\text{model}})$ tensor is then processed by the Mamba2 layer.\\
    
    \item Residual Connection: A residual or ``skip'' connection is applied around the normalization and Mamba2 layers. The original input to the block is added element-wise to the output of the Mamba2 layer.
\end{itemize}
After passing through the final Mamba block, the sequence is processed by one last RMSNorm layer. The resulting $(B, S, D_{\text{model}})$ tensor is then fed into the prediction head. This head is a single linear layer that projects the $D_{\text{model}}$-dimensional representation of each hit to a $3k$-dimensional vector, yielding a final output tensor of shape $(B, S, 3k)$. Here, $k=10$ is the number of neighbors to be predicted. This output format is designed specifically for the Causal k-Nearest Neighbor (kNN) objective.

\subsection{Maximal Update Parameterization}
\paragraph{\underline{Challenge in Scaling Models}} Imagine building with LEGOs. If you build a small car, it's stable. But if you try to build a life-sized car using the exact same small-brick techniques, it will be flimsy and fall apart. Modern AI models face a similar problem. When we try to make them bigger and more powerful by adding more ``width'' or digital neurons, their internal mathematics can become unstable during training. The signals inside can either ``explode'' into uselessly large numbers or ``vanish'' to zero, making it impossible for the model to learn. µ-Parameterization (µP) is a groundbreaking set of rules that solves this problem. It's like a master blueprint for building AI models, telling us exactly how to adjust the initial settings and the learning rate based on the model's size. This ensures that as the model scales up, its internal signals stay perfectly balanced, allowing it to train stably and effectively. A major benefit is that the best training settings found on a small, cheap model can be directly transferred to a massive, expensive one, saving enormous amounts of time and computational cost.

Concretely, standard infinite-width network analyses, such as those based on the Neural Tangent Kernel (NTK), predict that wide networks operate in a ``lazy regime'' where they fail to learn meaningful features from data. µ-Parameterization (µP) was introduced to overcome this limitation by defining a specific scaling of model initializations and learning rates that guarantees non-trivial feature evolution in the infinite-width limit. A significant practical advantage of µP is that it enables zero-shot hyperparameter transfer, allowing optimal settings found on small-scale models to be directly applied to their large-scale counterparts. This mitigates the often prohibitive computational costs associated with tuning large models.

\paragraph{\underline{Applications in Modern Architectures}}
The principles of µP have been successfully extended beyond simple multi-layer perceptrons (MLPs) to a range of complex architectures. In Transformers, µP facilitates hyperparameter transfer, although achieving a stable feature-learning limit requires careful scaling with respect to both model width and depth. The framework has also been adapted for scientific machine learning models like Fourier Neural Operators (FNOs), where a specific µ-FNO parameterization ensures stable training as the model size and number of Fourier modes are scaled. More recently, µP has been applied to stabilize the training of large Diffusion Models, again enabling hyperparameter transfer for these computationally intensive generative systems. This body of research highlights both the generality of the µP framework and the necessity of deriving architecture-specific scaling laws.

\paragraph{\underline{µP for MAMBA}}
To address this, a corrected scaling for State Space Models (SSMs), termed $\mu$P-SSM (Maximal Update Parameterization for SSMs), was derived by analyzing signal propagation directly within the Mamba architecture. This analysis yielded specific scaling rules for initialization variances ($\sigma$), which control the scale of the model's initial random weights, and learning rates ($\eta$), which determine the step size during training. The key formulas dictate how these parameters for Mamba's weight matrices ($W_B, W_C$) should be scaled relative to the model's latent state dimension ($N_x$) and input dimension ($N_u$). Using asymptotic Big-Theta ($\Theta$) notation, the rules are:
\begin{itemize}
    \item \textbf{Initialization Variances:} $\sigma_B \in \Theta(\sqrt{\frac{N_x}{N_u}})$ and $\sigma_C \in \Theta(\frac{1}{\sqrt{N_x N_u}})$
    \item \textbf{Learning Rates:} $\eta_B \in \Theta(\frac{N_x}{\sqrt{N_u}})$ and $\eta_C \in \Theta(\frac{1}{N_x \sqrt{N_u}})$
\end{itemize}
We have integrated this $\mu$P-SSM methodology into our own Mamba-based model. The effectiveness of this approach is evidenced by the stable scaling of layer-wise activation norms across different model sizes, as empirically verified in our experiments. Unlike standard parameterizations which lead to exploding signals or heuristic $\mu$P which leads to vanishing signals, our model's activations and their updates remain correctly scaled, confirming that the model is operating in a stable feature-learning regime.

\subsection{Additional Details for Pretraining}
The model is trained using the AdamW optimizer, which incorporates weight decay for regularization against overfitting. To manage the learning rate dynamics, we employ a cosine decay schedule, which is preceded by a brief linear warmup period at the beginning of training to ensure initial stability. To further prevent training instabilities arising from large gradients, we apply gradient clipping. The learning objective is to minimize a Mean Squared Error (MSE) loss function. This loss quantifies the Euclidean distance between the model's predicted coordinates for the k-Nearest Neighbors (kNN) and the truth coordinates. These truth neighbors are pre-computed for each particle spacepoint during the data loading phase to ensure efficient throughput during training.

\paragraph{\underline{Loss Re-scaling by Event Difficulty}}
We identified a nuisance structure in the training data related to event spacepoint density; events with a larger number of spacepoints are inherently easier to predict, as the average distance between neighboring spacepoints is smaller. This variance in difficulty can lead to training instability, manifesting as loss spikes. To mitigate this, we introduce a loss re-scaling strategy based on event binning. Events are first grouped into discrete bins based on their average k-Nearest Neighbor (kNN) distance, which serves as a proxy for prediction difficulty. Let $g(i)$ be the function that maps event $i$ to its corresponding difficulty bin. The loss objective is then modified as follows: (1) the Mean Squared Error (MSE) for each event is re-weighted by a factor $w_{g(i)}$ corresponding to the average difficulty of its bin, and (2) the total batch loss is calculated by averaging these re-weighted individual losses. This is formulated as:
\begin{equation*}
    \mathcal{L} = \frac{1}{B} \sum_{i=1}^{B} w_{g(i)} \mathcal{L}_i = \frac{1}{B} \sum_{i=1}^{B} w_{g(i)} \left( \frac{1}{S_n} \sum_{j=1}^{S_n} ||\mathbf{s}_{ij} - \mathbf{y}_{ij}||_2^2 \right)
\end{equation*}
Here, $B$ is the number of events in the batch, $\mathcal{L}_i$ is the standard MSE for event $i$ with $S_n$ spacepoints, $\mathbf{s}_{ij}$ and $\mathbf{y}_{ij}$ are the predicted and truth coordinates respectively, and $w_{g(i)}$ is the pre-computed weight for the difficulty bin to which the event belongs. This ensures that a single batch-averaged loss is computed only after accounting for the inherent difficulty of each event in the batch.

\section{Additional Results}
\label{app:results}
\subsection{Downstream Model}
\subsubsection{Tracking (Instance Segmentation)}

Our lightweight downstream model for track finding—formulated as a per-point instance segmentation task—is inspired by image panoptic segmentation models such as \textsc{MaskFormer} and \textsc{Mask2Former}, adapted to point cloud data.

Let $\mathbf{X} = \{ \mathbf{x}_i \}_{i=1}^N$ denote the input set of $N$ points, where each $\mathbf{x}_i \in \mathbb{R}^d$ is a $d$-dimensional point-level feature (either raw input, pretrained representation, or from a randomly initialized encoder). These are first projected into a latent embedding space via a linear layer:  
$$
\mathbf{e}_i = \mathbf{W}_\text{proj} \mathbf{x}_i, \quad \mathbf{e}_i \in \mathbb{R}^{d_e}.
$$
We denote the set of projected \pointembeds  as $\mathbf{E} = \{ \mathbf{e}_i \}_{i=1}^N$.

To represent candidate tracks, we use $K$ learnable queries (\tqueries) $\mathbf{Q}^{(0)} = \{ \mathbf{q}_k^{(0)} \}_{k=1}^K$, where each $\mathbf{q}_k^{(0)} \in \mathbb{R}^{d_e}$. These prototypes are refined over $L$ \td layers. Each decoder layer consists of:
\begin{itemize}
    \item \textbf{Cross-attention:} updates $\mathbf{q}_k$ by attending to point embeddings $\mathbf{E}$.
    \item \textbf{Self-attention:} refines interaction among the $K$ prototypes.
    \item \textbf{Feed-forward network (FFN):} standard transformer update.
\end{itemize}

After $L$ decoder layers, we obtain the \rtqueries $\mathbf{Q}^{(L)} = \{ \mathbf{q}_k^{(L)} \}_{k=1}^K$. Each refined query vector is then processed by two MLPs:
$$
\mathbf{m}_k = \text{MLP}_\text{mask}(\mathbf{q}_k^{(L)}), \quad
\hat{y}_k = \text{MLP}_\text{cls}(\mathbf{q}_k^{(L)}),
$$
where $\mathbf{m}_k \in \mathbb{R}^{d_e}$ is the \tembed for the $k$-th prototype, and \tinspred $\hat{y}_k \in [0,1]$ is the probability of corresponding to a real track (vs. a ``no-object'' class).

Each \tembed $\mathbf{m}_k$ is used to compute point-to-prototype assignment logits:
\[
z_{ik} = \mathbf{e}_i^\top \mathbf{m}_k, \quad \hat{p}_{ik} = \sigma(z_{ik}),
\]
where $\sigma(\cdot)$ denotes the sigmoid function. The predicted assignment probability $\hat{p}_{ik}$ represents the likelihood that point $i$ belongs to prototype $k$.

To encourage each \tquery to focus on the subset of points it is likely responsible for, we apply an \textit{additive attention mask} during cross-attention. The attention mask is defined as:
$$
A_{ik} = -\log(\hat{p}_{ik} + \epsilon),
$$
with a small constant $\epsilon$ added for numerical stability. This mask is added to the attention logits before the softmax operation in the cross-attention layer.
This dynamic masking suppresses contributions from low-probability points and improves localization by making each prototype attend selectively to its likely constituent points.

\paragraph{Training Loss.}  
Let $\mathcal{T} = \{ T_j \}_{j=1}^M$ be the set of $M$ ground-truth tracks (instance labels). We compute a bipartite matching between the $M$ ground-truth tracks and the $K$ \rtqueries using the Hungarian algorithm. The matching minimizes a cost function combining:
\begin{itemize}
\item Dice loss $\mathcal{L}_\text{dice}$ on the per-point predicted vs. ground-truth track,
\item Focal loss $\mathcal{L}_\text{focal}$ on point-wise assignment probabilities,
\item Classification loss $\mathcal{L}_\text{cls}$ on the track/no-object prediction.
\end{itemize}

For each matched pair $(T_j, \mathbf{q}_k)$, the total loss is:
\[
\mathcal{L}_\text{match}^{(j,k)} = \lambda_\text{dice} \cdot \mathcal{L}_\text{dice}^{(j,k)} + 
\lambda_\text{focal} \cdot \mathcal{L}_\text{focal}^{(j,k)} +
\lambda_\text{cls} \cdot \mathcal{L}_\text{cls}^{(k)}.
\]

For unmatched prototypes, we only compute $\mathcal{L}_\text{cls}^{(k)}$ with the ground truth label being ``no-object''.

The final training loss includes auxiliary losses from each decoder layer $\ell = 1, \dots, L$, as well as from the initial prototype vectors:
$$
\mathcal{L}_\text{total} = \sum_{\ell=0}^{L} \mathcal{L}^{(\ell)}.
$$
\noindent During inference, we assign each spacepoint $i$ to the track whose combined mask and classification score is maximal. Concretely, we compute
$
k_i^* = \arg\max_{k} \bigl(\hat p_{ik}\,\hat y_k\bigr),
$
and label point \(i\) as belonging to track \(k_i^*\).

\noindent This formulation enables end-to-end training of the instance segmentation model, while allowing the pretrained or learned point embeddings to guide track-level grouping.

\subsubsection{Particle Identification and Noise Identification}

For both PID and noise classification, we use a simple lightweight adapter:

\begin{itemize}
  \item \textbf{Embedding:} A linear layer projects each point feature $\mathbf{x}_i\in\mathbb{R}^d$ to a $d_p$-dimensional embedding.
  \item \textbf{Context:} A single Self-attention layer aggregates global information across all point embeddings.
  \item \textbf{Prediction:} An MLP with softmax over $C$ output classes.
\end{itemize}

\subsection{Comparative Methods for Downstream Tasks}
\label{app:comparative}

\subsubsection{Adapt \exatrkx Pipeline for \sphenix Tracking-Finding}
\label{app:exatrkx}

In this section, we discuss the several adaptions to the \exatrkx pipeline for it to work well on the \sphenix data. We need to apply adaptions to the first four stages -- data pre-processing, hit embedding, edge filtering, and GNN edge classification -- out of six stages of the \exatrkx pipeline.

\noindent\textbf{Pre-processing.} The \exatrkx's study was based on the \trackml dataset~\cite{m:trackml}. The dataset provides two sources for the construction of the neural network input -- the $3$ dimensional location of the spacepoints and the directional information and summary statistics from the charge deposited in each spacepoint ($8$-dimensional). The second source of information is called cell features by the paper. The hit feature is the concatenation of the location and cell features. 
Since \sphenix data does not provide cell features, we only used the location of hits in the HEP-coordinate to construct the input. More precisely, let $(\hat{\eta}, \phi, \hat{r})$ be the location of a hit (normalized pseudorapidity, angle, and normalized radius), the features of this hit is a $5$-dimensional vector
\[
    \paren{\mathcal{E}, \hat{\eta}, \cos(\phi), \sin(\phi), \hat{r}},
\]
where $\mathcal{E}$ is the energy. We used $(\cos(\phi), \sin{\phi})$ instead of $\phi$ to overcome the discontinuity of $\phi$ at $2\pi$. 
We normalized the pseudorapidity $\eta$ by $1.96$ to get the normalized pseudorapidity $\hat{\eta}\in(-1, 1)$. To normalized a radius, we first match it to the closest one of the $48$ radius bins and use the bin index to replace the radius. And then, we divided the index by $48$ to normalized radius to a number between $[0, 1)$. We do this because the distance between the \sphenix TPC layers are not uniform with outer layers spacing farther apart than the inner ones. This may be a problem for distance-based edge set construction for a GNN model since same-track hits toward the end of the track may be less likely to be connected by the model.   

\vspace{2pt}
\noindent\textbf{Embedding and filtering.} The \exatrkx pipeline embeds the spacepoints and filters the edges as two separate steps. To adapt them for \sphenix, we modified the procedure in the following aspects: 1) how to determine whether a pair of hits is connected; 2) how candidate hit pairs are generated; 3) how to trained the models; and 4) how to construct the models.

In the embedding stage, \exatrkx trained a multi-layer perceptron (MLP) network to embed each hit into a latent representation so that pairs of neighboring hits from the same track are closer in the latent space than pairs that are not (e.g.~from different tracks or not neighbors on the same track). The embedding network is trained by first passing the two hits through the same embedding network and then minimizing the hinge loss of the distance between the two embeddings. 

Since \sphenix data does not provide information to determine whether two same-track hits are direct neighbors (although this information could be inferred for high-energy tracks), we decided not to distinguish whether two same-track hits are neighboring or not. This approach was also recommended by the \exatrkx research as a valid alternative. 

In the filtering stage, \exatrkx takes a pair of hits, passes them through the embedding network, concatenates the two embeddings, and pass the concatenation through a MLP filtering network to predict whether the two hits are connected. The prediction is optimized by a binary cross entropy loss. 

For both the embedding and filtering models, we need to provide candidate hit pairs. For the embedding stage, \exatrkx uses two types of pairs: random pairs and k-nearest neighbor (KNN) pairs as a form of hard negative mining. As a random pair has an extremely low chance to be connected, \exatrkx also trains on pairs formed by a hit and its closest neighbors in the latent representation space.

We follow the pipeline as closely as possible. However, because of the different between \sphenix and \trackml input features and the fact we treating all pairs from the same track as being connected (in contrast to \exatrkx's approach where only immediate neighbors are connected), we had to choose different cutoffs in both embedding and filtering. More specifically, we set a threshold of $2.$ for distance in the embedding space with pairs less than the threshold apart classified as having an edge between them. The threshold was so chosen as it ensures that we have an over $.8$ recall (efficiency in the \exatrkx terminology) in identifying pairs from the same track. Note here we didn't selected a threshold that will ensure close to a $100\%$ recall. This is because we can afford the model to fail to recognize faraway points from the same track as being connected. 

For the filtering step, we chose a threshold of $.675$ for probability of a true edge with pairs over the threshold considered as being connected. The threshold was selected because it ensures the false positive rate in edge identification to go below $1\%$. 

\vspace{2pt}
\noindent\textbf{GNN edge classification.}
For the final GNN step, we also used the Interaction Network~\cite{battaglia2016interaction} architecture with the same hyperparameters used by the \exatrkx pipeline. For edge classification, we chose a threshold of $.9$ as probability of a true edge. With this choice, we achieved a $91.79\%$ tracking efficiency (recall) (and $94.74\%$ for tracks with $\pt > \gev{1}$), and a track purity (precision) of $66.42\%$. With a threshold of $.8$, the tracking efficiency drops slightly to $90.01\%$ (and $92.60\%$ for $\pt > \gev{1}$) with a large improvement in purity to $76.72\%$.

\subsubsection{Adapt \eggnet for \sphenix Track-Finding}
\label{app:adaption_of_the_eggnet}
The \eggnet study was also based on the \trackml dataset~\cite{m:trackml} and share the same data pre-processing approach with \exatrkx. To partially compensate the lack cell features from \sphenix data, we tried the following approach to augmented the input. Let $\paren{\hat{\eta}_0, \phi_0, \hat{r}_0}$ be the location of a hit (normalized pseudorapidity, angle, and normalized radius), the features of this hit is a $12$-dimensional vector
\[
    \paren{\hat{\eta}_0, \cos(\phi_0), \sin(\phi_0), \hat{r}_0;\, \hat{\eta}_1, \cos(\phi_1), \sin(\phi_1), \hat{r}_1;\, \hat{\eta}_2, \cos(\phi_2), \sin(\phi_2), \hat{r}_2},
\]
where $\paren{\hat{\eta}_1, \phi_1, \hat{r}_1}$ and $\paren{\hat{\eta}_2, \phi_2, \hat{r}_2}$ are the locations of the two closest neighbors of the hit in the $\paren{\hat{\eta}, \cos(\phi), \sin(\phi), \hat{r}}$ space. The motivation for augmenting the hit with two closest neighbors is that for the majority of the hits in a high energy track, the two closest neighbors are most likely from the same track in which case the augmented hit features can provide information on the direction of track.

For the GNN model, \eggnet adopted a similar approach to \gravnet~\cite{qasim2019learning}. The outstanding feature of a \gravnet-type model is that the edge set is not predetermined but constructed dynamically. More precisely, \eggnet will run $N$ normal GNN iterations, but before each GNN iteration, the edge set will be constructed via KNN based on the current node embeddings. To adapt \eggnet to \sphenix data, we set GNN iterations to be $4$ and used $4$ message-passing rounds for each GNN iteration. The nearest $10$ hits in the embedding space are used to form the neighborhood of a hit. Different from the original \gravnet (but similar to the interaction GNN used by \exatrkx), \eggnet also has an edge network for calculating edge messages. Moreover, \eggnet also used a dedicated node decoding network to produce the node embeddings for the KNN. All sub-networks of \eggnet (node encoding/decoding networks and edge network) are MLPs with $2$ hidden layers and $64$ hidden feature each. The embedding dimension of the node (i.e.~the number output features node decoding network) is $24$. 

The network is trained with a hinge loss of margin $1$, aiming at reducing the Euclidean distance in the embedding space between a pair of hits from the same track and enlarge the distance between a pair from different tracks. The model was trained for $300$ epochs and the final clustering was done using DBSCAN with $\epsilon=1$ and minimum number samples $=2$. 

\subsubsection{Adapt GNNs for \sphenix Particle Identification and Noise Tagging}
\label{app:gnn}
We selected four GNN models: \GATConv, \GCNConv, \GraphConv, and \SAGEConv as benchmarking algorithms for the PID and noise-tagging downstream tasks.
We used the \texttt{torch\_geometric} implementations for the models. We used the same data pre-processing protocol as discussed in~\ref{app:exatrkx}.
To generate the edge set, for a hit at location $\paren{\eta, \cos(\phi), \sin(\phi), \hat{r}}$, we connect to it $50$ nearest neighbor hits with distance $<1$. We allowed the edges to be directed. The node features to the GNNs are the energy $\mathcal{E}$ of the hit together with its 4D location. For the node encoding network, we use a MLP with $2$ hidden layers and $256$ hidden features each. We use uniformly $6$ GNN layers for each GNN model. For the hit classification network, we use a MLP of $2$ hidden layers with $128$ and $64$ hidden features. The GNNs are trained with cross entropy loss. Each GNN is trained for $200$ epochs. 

In general, GNNs' performance on the two downstream tasks are suboptimal. We hypothesis that the failure of GNNs is a result of their difficulty in capturing and communicating more global patterns of the tracks as solving both particle identification and noise-tagging require a model to understand the general shape of tracks that span a significant space in TPC. 

\subsubsection{Adapt \oneformer for \sphenix Particle Identification and Noise Tagging}
\label{app:oneformer}
\oneformer is a state-of-the-art object detection algorithm for 3D point cloud data that can solve semantic and instance segmentation task in one run. The model architecture of \oneformer is U-Net backbone followed by a Transformer decoder. 

To run \oneformer on a point cloud data, we first need to get the so-called super points (a grouping of raw points) either by a clustering algorithm or voxelization. To adapt \oneformer to \sphenix data, we used the same pre-processing approach as discussed in \ref{app:exatrkx} and voxelized the resulting point cloud to a grid of shape $(64, 64, 48)$ in $\hat{\eta}, \phi, \hat{r}$, respectively. 

The super points first pass through the sparse convolution-powered U-Net backbone to be featurized. Then the super point features serve as the keys and values in the Transformer encoder. The learnable queries output from the Transformer decoder are then used to produce instance/semantic segmentation predictions on the super points. In the final step, the prediction on the super points will be broadcast to their constituent raw points. Since both particle identification and noise-tagging can be considered as semantic segmentation tasks, we separated the part of the code (primarily in prediction and loss function) for semantic segmentation from \oneformer, while kept the neural architecture identical. We used the same network parameters as the example of \oneformer on the S3DIS dataset. 

\begin{figure}[ht]
    \centering
    \includegraphics[width=1\textwidth]{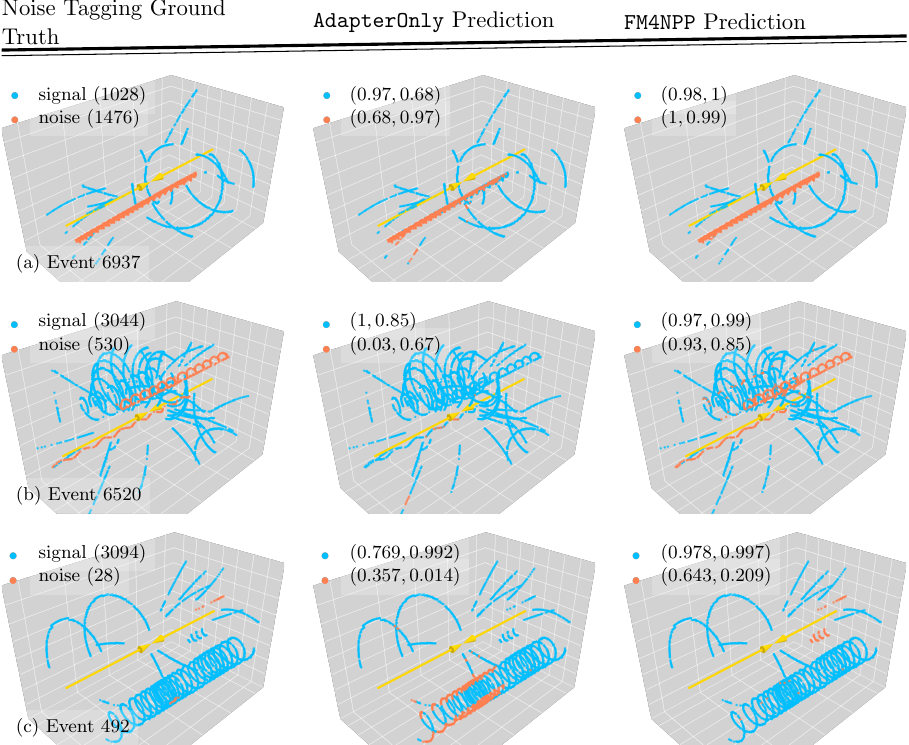}
    \caption{\textbf{Performance of \fmhead and \fm on Noise Tagging.} The numbers in the parentheses in the target sub-figures are the number of signal and noise spacepoints. The numbers in the parentheses in the prediction sub-figures are the recall and precision of the class.}
    \label{fig:app_noise_performance}
\end{figure}

\begin{figure}[ht]
    \centering
    \includegraphics[width=1\textwidth]{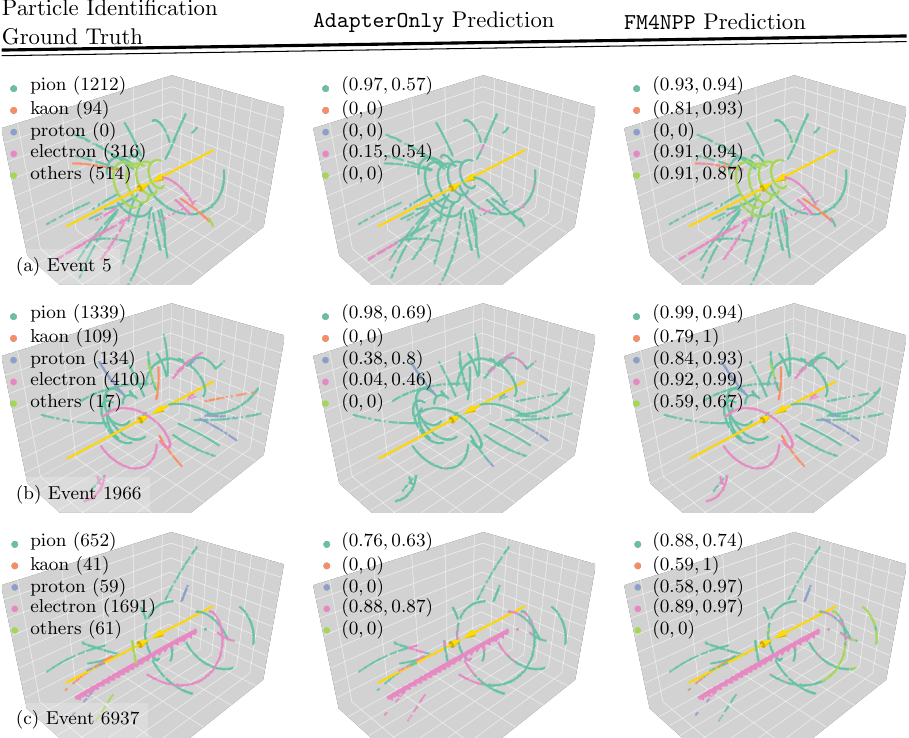}
    \caption{\textbf{Performance of \fmhead and \fm on particle identification.} The numbers in the parentheses in the target sub-figures are the number of spacepoints in each particle ID class. The numbers in the parentheses in the prediction sub-figures are the recall and precision of the class.}
    \label{fig:app_noise_performance}
\end{figure}

\begin{figure}[ht]
    \centering
    \includegraphics[width=1\textwidth]{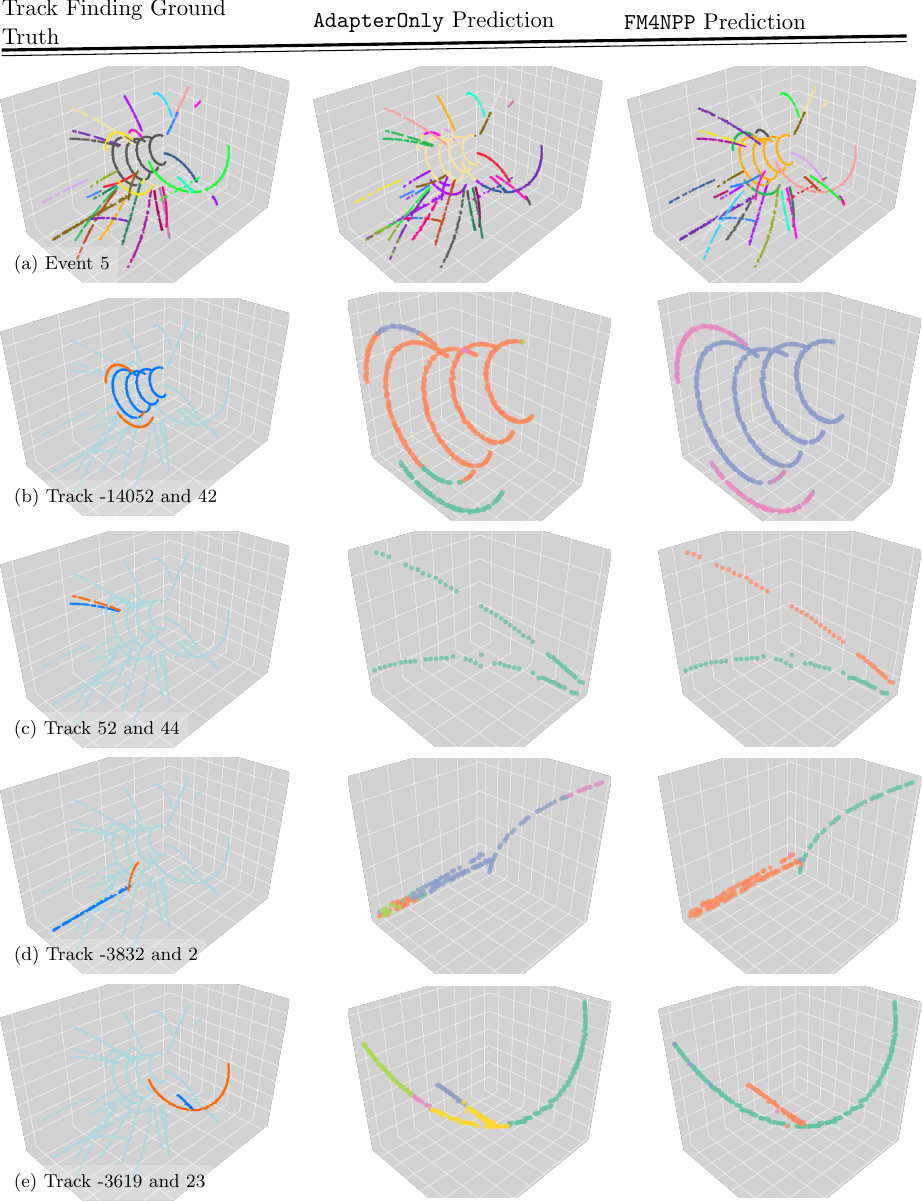}
    \caption{\textbf{Performance of \fmhead and \fm on track finding.} In panel (a), we show the ground-truth tracks, the \fmhead track candidates, and the \fm track candidates (note that two different tracks might have the same color since the length of the color cycle we used may be smaller than the number of tracks). In panel (b)-(e), we show four pairs of close-by ground-truth tracks that the \fmhead model fails to separate while the \fm model does.}
    \label{fig:app_noise_performance}
\end{figure}

\begin{table*}[ht]
\centering
\begin{tabular}{l c cc cc cc}
\toprule
\rule{0pt}{12pt} & Accuracy 
  & \multicolumn{2}{c}{Macro} 
  & \multicolumn{2}{c}{Non‑noise} 
  & \multicolumn{2}{c}{Noise} \\
\cmidrule(lr){3-4} \cmidrule(lr){5-6} \cmidrule(lr){7-8}
      &          
  & Recall & Precision 
  & Recall & Precision 
  & Recall & Precision \\
\midrule
\GATConv   & $0.9099$ & $0.6730$ & $0.8060$ & $0.9788$ & $0.9242$ & $0.3672$ & $0.6878$ \\
\GCNConv   & $0.9095$ & $0.6728$ & $0.8037$ & $0.9784$ & $0.9241$ & $0.3672$ & $0.6832$ \\
\GraphConv & $0.9190$ & $0.7213$ & $0.8252$ & $0.9764$ & $0.9351$ & $0.4661$ & $0.7152$ \\
\SAGEConv  & $0.9174$ & $0.7227$ & $0.8165$ & $0.9740$ & $0.9355$ & $0.4714$ & $0.6975$ \\
\oneformer & $0.9646$ & $0.9404$ & $0.8948$ & $0.9716$ & $0.9884$ & $0.9092$ & $0.8012$ \\
\fmhead    & $0.9111$ & $0.6215$ & $0.8359$ & $0.9901$ & $0.9169$ & $0.2528$ & $0.7548$ \\
\fmmvi        & $0.9662$ & $0.9122$ & $0.9114$ & $0.9809$ & $0.9812$ & $0.8435$ & $0.8416$ \\

\bottomrule
\end{tabular}
\caption{Noise tagging per-class recall and precision.}
\label{tab:noise-tagging-full}
\end{table*}

\begin{table*}[ht]
\centering
\caption{Particle Identification per-class recall and precision.}
\resizebox{\linewidth}{!}{
\setlength{\tabcolsep}{4pt}
\begin{tabular}{l c c *{6}{cc}}
\toprule
 \rule{0pt}{12pt} & Accuracy
  & \multicolumn{2}{c}{Macro}
  & \multicolumn{2}{c}{Others}
  & \multicolumn{2}{c}{Pion}
  & \multicolumn{2}{c}{Kaon}
  & \multicolumn{2}{c}{Proton}
  & \multicolumn{2}{c}{Electron} \\
\cmidrule(lr){3-4}
\cmidrule(lr){5-6}
\cmidrule(lr){7-8}
\cmidrule(lr){9-10}
\cmidrule(lr){11-12}
\cmidrule(lr){13-14}
  & 
  & Rec. & Pre. 
  & Rec. & Pre. 
  & Rec. & Pre. 
  & Rec. & Pre. 
  & Rec. & Pre. 
  & Rec. & Pre. \\
\midrule
\GATConv   & $0.6922$ & $0.3973$ & $0.6368$ & $0.0947$ & $0.5709$ & $0.9106$ & $0.7014$ & $0.0057$ & $0.6146$ & $0.4567$ & $0.6117$ & $0.5190$ & $0.6854$ \\
\GCNConv   & $0.6892$ & $0.3911$ & $0.6319$ & $0.0782$ & $0.5762$ & $0.9140$ & $0.6966$ & $0.0073$ & $0.5871$ & $0.4501$ & $0.6140$ & $0.5058$ & $0.6858$ \\
\GraphConv & $0.7079$ & $0.4176$ & $0.6425$ & $0.1304$ & $0.5739$ & $0.9133$ & $0.7146$ & $0.0080$ & $0.5791$ & $0.4766$ & $0.6272$ & $0.5597$ & $0.7178$ \\
\SAGEConv  & $0.7262$ & $0.4563$ & $0.6502$ & $0.1085$ & $0.5790$ & $0.9126$ & $0.7382$ & $0.0338$ & $0.5239$ & $0.6242$ & $0.7071$ & $0.6024$ & $0.7028$ \\
\oneformer & $0.7701$ & $0.4897$ & $0.5767$ & $0.3029$ & $0.5758$ & $0.9207$ & $0.7658$ & $0.0000$ & $0.0000$ & $0.4859$ & $0.6991$ & $0.7389$ & $0.8427$ \\
\fmhead    & $0.6631$ & $0.3387$ & $0.6111$ & $0.0095$ & $0.7714$ & $0.9511$ & $0.6596$ & $0.0002$ & $0.2872$ & $0.4120$ & $0.6366$ & $0.3209$ & $0.7008$ \\
\fmmvi        & $0.8547$ & $0.6623$ & $0.8328$ & $0.4449$ & $0.7647$ & $0.9551$ & $0.8484$ & $0.2712$ & $0.7829$ & $0.8068$ & $0.8763$ & $0.8336$ & $0.8919$ \\
\bottomrule
\end{tabular}
}
\label{tab:pid_metrics_full}
\end{table*}

\begin{table*}[ht]
    \centering
    \caption{\textbf{Diagnostic metrics for tracking performance.}}
    \begin{tabular}{lcccc}
    \toprule
    \rule{0pt}{12pt}model & ARI & overall spacepoint efficiency & overall spacepoint purity & no.~parameters \\
    \midrule
    
    \eggnet  & $0.7256$ & $93.01\%$ & $92.34\%$ & $0.16$M \\
    \exatrkx & $0.8765$ & $94.47\%$ & $98.83\%$ & $3.86$M \\
    \fmhead  & $0.7243$ & $89.34\%$ & $92.09\%$ & $2.39$M \\ 
    \fmmvi     & $0.9448$ & $97.56\%$ & $98.34\%$ & $188$M + $2.39$M \\
    \bottomrule
    \end{tabular}
    \label{tab:performance}
\end{table*}

\subsection{Additional Learned Embeddings Results}
\label{app:xai}

\begin{figure*}[ht]
\begin{center}
\includegraphics[width=\linewidth]{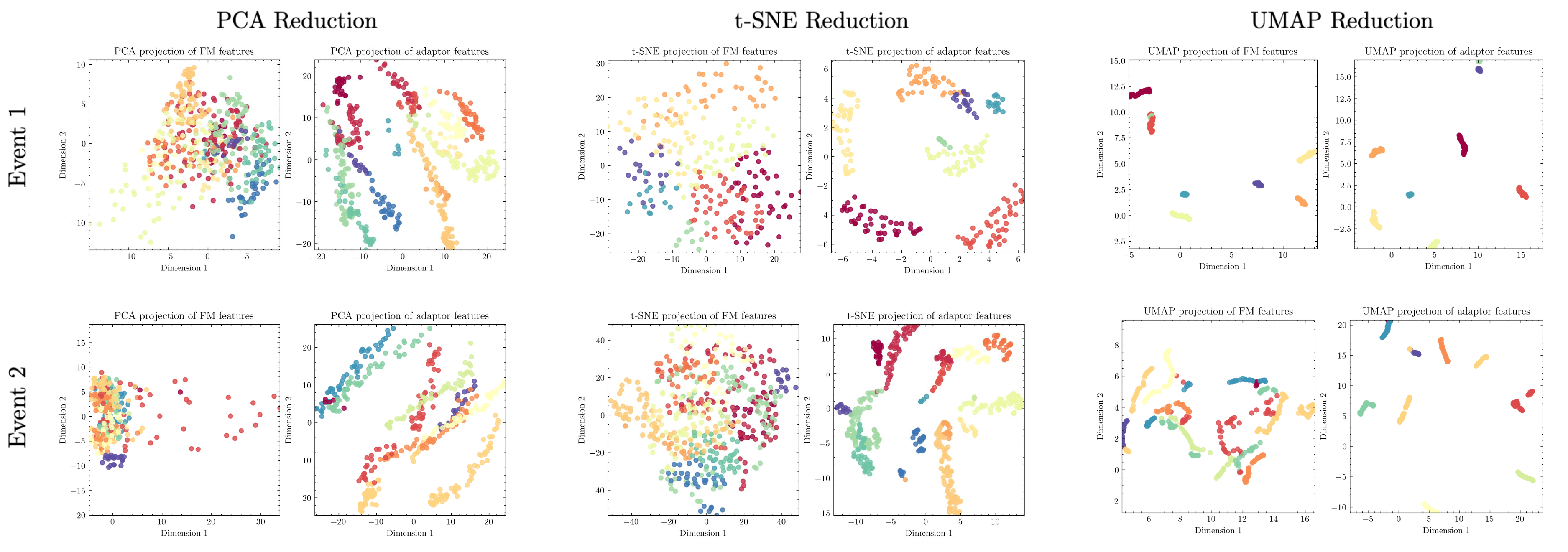}
\caption{Dimensionality reduction results using PCA, t-SNE, and UMAP on randomly selected test data samples.}
\label{fig:different_reduction_methods}
\end{center}
\end{figure*}

To investigate and interpret the effectiveness of learned latent representations from our pre-trained foundation model (FM), we conducted a comprehensive explainable AI (XAI) analysis. In Figure~\ref{fig:different_reduction_methods}, we present results obtained by applying various dimensionality reduction techniques (including PCA, t-SNE and UMAP) to both FM features and downstream adaptor features. For illustrative clarity, we randomly selected two test data samples. The results demonstrate consistent improvement, clearly showcasing the FM features' adaptability: even after a single linear projection, the FM embeddings exhibit substantial clustering and separability, indicating rapid adaptation to the downstream track-finding task. Adaptor features consistently provided superior discrimination, yielding distinctly well-separated clusters corresponding to different track categories.

\begin{figure*}[ht]
\begin{center}
\includegraphics[width=\linewidth]{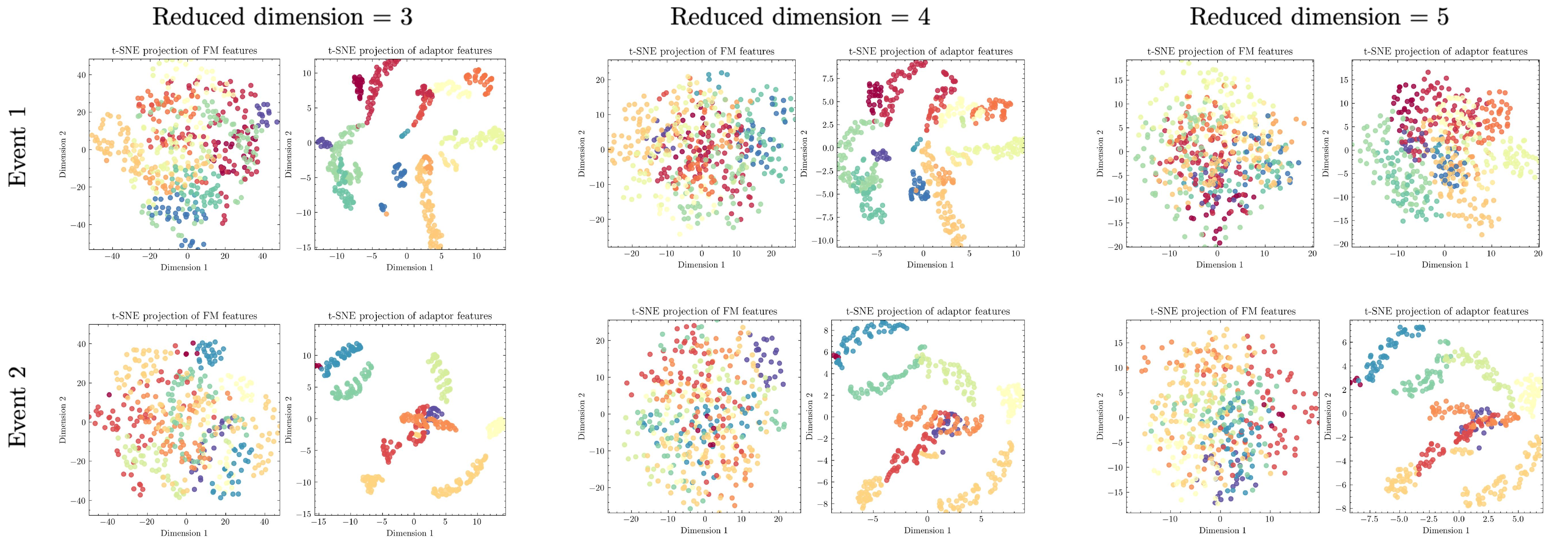}
\caption{T-SNE visualizations for randomly selected test instances across various reduced dimensions.}
\label{fig:different_reduced_dimensions}
\end{center}
\end{figure*}

To further validate the robustness and generalizability of the FM features, we systematically investigated the impact of varying dimensionality reduction parameters using t-SNE. Specifically, we conducted experiments by setting the reduced dimensionality to 3, 4, and 5 and visualized the results by plotting the first two t-SNE components (See Figure~\ref{fig:different_reduced_dimensions}). Across all tested dimensional configurations, the FM features consistently demonstrated pronounced clustering patterns and clear separability, highlighting their intrinsic adaptability and effectiveness in supporting diverse downstream classification tasks.

\begin{figure*}[ht]
\begin{center}
\includegraphics[width=\linewidth]{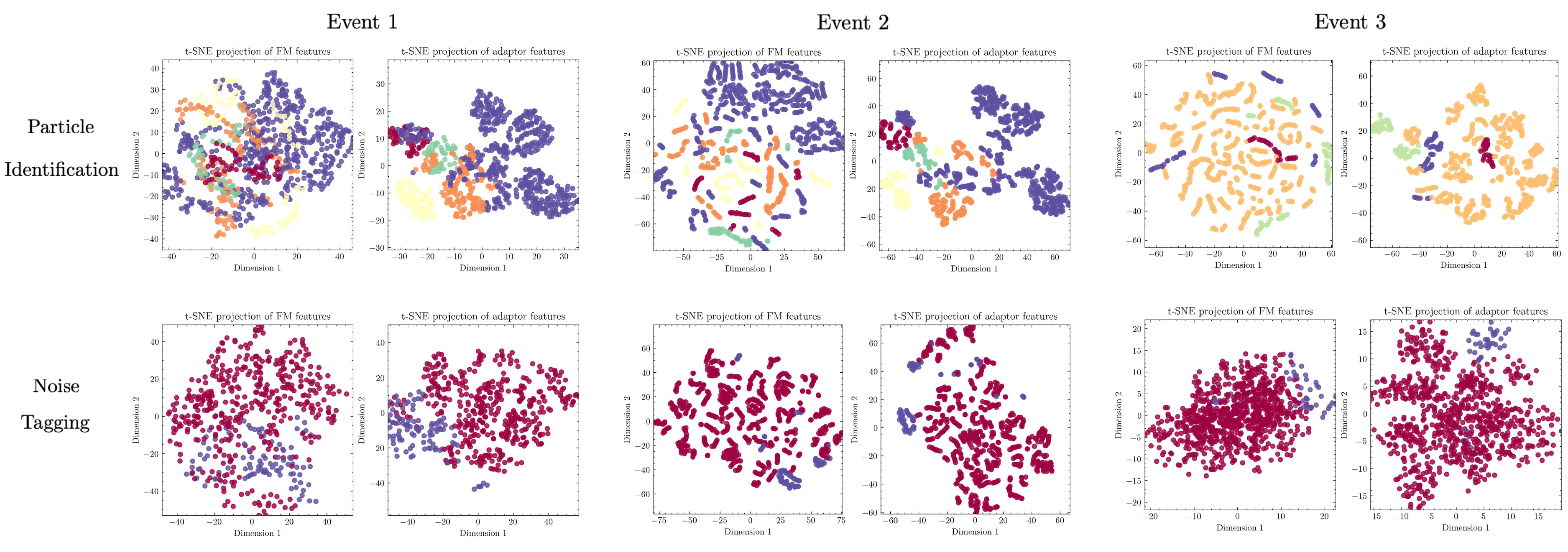}
\caption{T-SNE visualizations for randomly selected test instances across various downstream tasks.}
\label{fig:different_downstream_tasks}
\end{center}
\end{figure*}

In Figure~\ref{fig:different_downstream_tasks}, we extended our analysis to multiple downstream tasks, again using randomly selected test data instances and employing t-SNE for visualization. The FM features' separability was notably effective for the track-finding task, slightly diminished for particle identification, and considerably reduced for noise tagging. The limited performance observed in noise tagging is attributed to the inherent imbalance of the binary classification data, making separability challenging due to the dominant prevalence of a single label. Overall, our analyses confirm a hierarchy of effectiveness in FM embeddings across downstream tasks: track-finding demonstrates the strongest separability, followed by particle identification, and lastly noise tagging. These findings align well with the FM's pretraining objective, neighbor identification, and are consistent with task relevance from a physics perspective.

\clearpage

\bibliography{bib}

\begin{thebibliography}{51}
\providecommand{\natexlab}[1]{#1}

\bibitem[{Agostinelli et~al.(2003)}]{m:g4}
Agostinelli, S.; et~al. 2003.
\newblock {GEANT4 - A Simulation Toolkit}.
\newblock \emph{Nucl. Instrum. Meth. A}, 506: 250--303.

\bibitem[{Aguilar et~al.(2022)Aguilar, Chang, Elayavalli, Fatemi, He, Ji,
  Kalinkin, Kelsey, Mooney, and Verkest}]{PYTHIA_tune}
Aguilar, M.~R.; Chang, Z.; Elayavalli, R.~K.; Fatemi, R.; He, Y.; Ji, Y.;
  Kalinkin, D.; Kelsey, M.; Mooney, I.; and Verkest, V. 2022.
\newblock {pythia8 underlying event tune for RHIC energies}.
\newblock \emph{Phys. Rev. D}, 105(1): 016011.

\bibitem[{Amrouche et~al.(2020)Amrouche, Basara, Calafiura, Estrade, Farrell,
  Ferreira, Finnie, Finnie, Germain, Gligorov, Golling, Gorbunov, Gray, Guyon,
  Hushchyn, Innocente, Kiehn, Moyse, Puget, Reina, Rousseau, Salzburger,
  Ustyuzhanin, Vlimant, Wind, Xylouris, and Yilmaz}]{m:trackml}
Amrouche, S.; Basara, L.; Calafiura, P.; Estrade, V.; Farrell, S.; Ferreira,
  D.~R.; Finnie, L.; Finnie, N.; Germain, C.; Gligorov, V.~V.; Golling, T.;
  Gorbunov, S.; Gray, H.; Guyon, I.; Hushchyn, M.; Innocente, V.; Kiehn, M.;
  Moyse, E.; Puget, J.-F.; Reina, Y.; Rousseau, D.; Salzburger, A.;
  Ustyuzhanin, A.; Vlimant, J.-R.; Wind, J.~S.; Xylouris, T.; and Yilmaz, Y.
  2020.
\newblock The Tracking Machine Learning Challenge: Accuracy Phase.
\newblock In Escalera, S.; and Herbrich, R., eds., \emph{The NeurIPS '18
  Competition}, 231--264. Cham: Springer International Publishing.

\bibitem[{Battaglia et~al.(2016)Battaglia, Pascanu, Lai, Jimenez~Rezende
  et~al.}]{battaglia2016interaction}
Battaglia, P.; Pascanu, R.; Lai, M.; Jimenez~Rezende, D.; et~al. 2016.
\newblock Interaction networks for learning about objects, relations and
  physics.
\newblock \emph{Advances in neural information processing systems}, 29.

\bibitem[{Belmont et~al.(2024)}]{Belmont:2023fau}
Belmont, R.; et~al. 2024.
\newblock {Predictions for the sPHENIX physics program}.
\newblock \emph{Nucl. Phys. A}, 1043: 122821.

\bibitem[{Bodnar et~al.(2025)Bodnar, Bruinsma, Lucic, Stanley, Allen,
  Brandstetter, Garvan, Riechert, Weyn, Dong et~al.}]{aurora}
Bodnar, C.; Bruinsma, W.~P.; Lucic, A.; Stanley, M.; Allen, A.; Brandstetter,
  J.; Garvan, P.; Riechert, M.; Weyn, J.~A.; Dong, H.; et~al. 2025.
\newblock A foundation model for the Earth system.
\newblock \emph{Nature}, 1--8.

\bibitem[{Bommasani et~al.(2021)Bommasani, Hudson, Adeli, Altman, Arora, von
  Arx, Bernstein, Bohg, Bosselut, Brunskill
  et~al.}]{bommasani2021opportunities}
Bommasani, R.; Hudson, D.~A.; Adeli, E.; Altman, R.; Arora, S.; von Arx, S.;
  Bernstein, M.~S.; Bohg, J.; Bosselut, A.; Brunskill, E.; et~al. 2021.
\newblock On the opportunities and risks of foundation models.
\newblock \emph{arXiv preprint arXiv:2108.07258}.

\bibitem[{{Brookhaven National Laboratory}(2025)}]{sphenix_detector_bnl}
{Brookhaven National Laboratory}. 2025.
\newblock {sPHENIX Detector at RHIC}.
\newblock \url{https://www.bnl.gov/rhic/sphenix.php}.
\newblock Accessed: 2025-07-19.

\bibitem[{Busza, Rajagopal, and van~der Schee(2018)}]{m:HIintro}
Busza, W.; Rajagopal, K.; and van~der Schee, W. 2018.
\newblock {Heavy Ion Collisions: The Big Picture, and the Big Questions}.
\newblock \emph{Ann. Rev. Nucl. Part. Sci.}, 68: 339--376.

\bibitem[{Calafiura et~al.(2024)Calafiura, Chan, Delabrouille, and
  Wang}]{rw:EggNet}
Calafiura, P.; Chan, J.; Delabrouille, L.; and Wang, B. 2024.
\newblock EggNet: An Evolving Graph-based Graph Attention Network for Particle
  Track Reconstruction.
\newblock arXiv:2407.13925.

\bibitem[{Calafiura et~al.(2018)Calafiura, Farrell, Gray, Vlimant, Innocente,
  Salzburger, Amrouche, Golling, Kiehn, Estrade, Germaint, Guyon, Moyse,
  Rousseau, Yilmaz, Gligorov, Hushchyn, and Ustyuzhanin}]{trackml2018}
Calafiura, P.; Farrell, S.; Gray, H.; Vlimant, J.-R.; Innocente, V.;
  Salzburger, A.; Amrouche, S.; Golling, T.; Kiehn, M.; Estrade, V.; Germaint,
  C.; Guyon, I.; Moyse, E.; Rousseau, D.; Yilmaz, Y.; Gligorov, V.~V.;
  Hushchyn, M.; and Ustyuzhanin, A. 2018.
\newblock TrackML: A High Energy Physics Particle Tracking Challenge.
\newblock In \emph{2018 IEEE 14th International Conference on e-Science
  (e-Science)}, 344--344.

\bibitem[{Cheng et~al.(2022)Cheng, Misra, Schwing, Kirillov, and
  Girdhar}]{rw:mask2former}
Cheng, B.; Misra, I.; Schwing, A.~G.; Kirillov, A.; and Girdhar, R. 2022.
\newblock Masked-attention Mask Transformer for Universal Image Segmentation.
\newblock arXiv:2112.01527.

\bibitem[{Cheng, Schwing, and Kirillov(2021)}]{rw:maskformer}
Cheng, B.; Schwing, A.~G.; and Kirillov, A. 2021.
\newblock Per-Pixel Classification is Not All You Need for Semantic
  Segmentation.
\newblock arXiv:2107.06278.

\bibitem[{Collaboration, Aad et~al.(2012)}]{aad2012observation}
Collaboration, A.; Aad, G.; et~al. 2012.
\newblock Observation of a New Particle in the Search for the Standard Model
  Higgs Boson with the ATLAS Detector at the LHC.
\newblock \emph{Physics Letters B}, 716(1): 1--29.

\bibitem[{Dao and Gu(2024)}]{mamba2}
Dao, T.; and Gu, A. 2024.
\newblock Transformers are {SSM}s: Generalized Models and Efficient Algorithms
  Through Structured State Space Duality.
\newblock In \emph{Forty-first International Conference on Machine Learning}.

\bibitem[{Dosovitskiy et~al.(2021)Dosovitskiy, Beyer, Kolesnikov, Weissenborn,
  Zhai, Unterthiner, Dehghani, Minderer, Heigold, Gelly, Uszkoreit, and
  Houlsby}]{rw5}
Dosovitskiy, A.; Beyer, L.; Kolesnikov, A.; Weissenborn, D.; Zhai, X.;
  Unterthiner, T.; Dehghani, M.; Minderer, M.; Heigold, G.; Gelly, S.;
  Uszkoreit, J.; and Houlsby, N. 2021.
\newblock An Image is Worth 16x16 Words: Transformers for Image Recognition at
  Scale.
\newblock In \emph{International Conference on Learning Representations}.

\bibitem[{Fedus, Zoph, and Shazeer(2022)}]{fedus2022switch}
Fedus, W.; Zoph, B.; and Shazeer, N. 2022.
\newblock Switch transformers: Scaling to trillion parameter models with simple
  and efficient sparsity.
\newblock \emph{Journal of Machine Learning Research}, 23(120): 1--39.

\bibitem[{Gu and Dao(2023)}]{mamba}
Gu, A.; and Dao, T. 2023.
\newblock Mamba: Linear-time sequence modeling with selective state spaces.
\newblock \emph{arXiv preprint arXiv:2312.00752}.

\bibitem[{Gu and Dao(2024)}]{rw6}
Gu, A.; and Dao, T. 2024.
\newblock Mamba: Linear-Time Sequence Modeling with Selective State Spaces.
\newblock In \emph{First Conference on Language Modeling}.

\bibitem[{Hoffmann et~al.(2022)Hoffmann, Borgeaud, Mensch, Buchatskaya, Cai,
  Rutherford, Casas, Hendricks, Welbl, Clark et~al.}]{nlpscale1}
Hoffmann, J.; Borgeaud, S.; Mensch, A.; Buchatskaya, E.; Cai, T.; Rutherford,
  E.; Casas, D. d.~L.; Hendricks, L.~A.; Welbl, J.; Clark, A.; et~al. 2022.
\newblock Training compute-optimal large language models.
\newblock \emph{arXiv preprint arXiv:2203.15556}.

\bibitem[{Hubert and Arabie(1985)}]{e:ari}
Hubert, L.; and Arabie, P. 1985.
\newblock Comparing partitions.
\newblock \emph{Journal of Classification}, 2(1): 193--218.

\bibitem[{Jiang and Qian(2025)}]{rw:trackingssm}
Jiang, C.; and Qian, S. 2025.
\newblock Application of Structured State Space Models to High energy physics
  with locality sensitive hashing.
\newblock In \emph{International Conference on Artificial Intelligence and
  Statistics}, 3961--3969. PMLR.

\bibitem[{Ju et~al.(2021)Ju, Murnane, Calafiura, Choma, Conlon, Farrell, Xu,
  Spiropulu, Vlimant, Aurisano, Hewes, Cerati, Gray, Klijnsma, Kowalkowski,
  Atkinson, Neubauer, DeZoort, Thais, Chauhan, Schuy, Hsu, Ballow, and
  Lazar}]{rw:Exatrack1}
Ju, X.; Murnane, D.; Calafiura, P.; Choma, N.; Conlon, S.; Farrell, S.; Xu, Y.;
  Spiropulu, M.; Vlimant, J.-R.; Aurisano, A.; Hewes, J.; Cerati, G.; Gray, L.;
  Klijnsma, T.; Kowalkowski, J.; Atkinson, M.; Neubauer, M.; DeZoort, G.;
  Thais, S.; Chauhan, A.; Schuy, A.; Hsu, S.-C.; Ballow, A.; and Lazar, A.
  2021.
\newblock Performance of a geometric deep learning pipeline for HL-LHC particle
  tracking.
\newblock \emph{The European Physical Journal C}, 81(10).

\bibitem[{Kalman(1960)}]{rw:kf}
Kalman, R.~E. 1960.
\newblock A New Approach to Linear Filtering and Prediction Problems.
\newblock \emph{Journal of Basic Engineering}, 82(1): 35--45.

\bibitem[{Kaplan et~al.(2020)Kaplan, McCandlish, Henighan, Brown, Chess, Child,
  Gray, Radford, Wu, and Amodei}]{nlpscale2}
Kaplan, J.; McCandlish, S.; Henighan, T.; Brown, T.~B.; Chess, B.; Child, R.;
  Gray, S.; Radford, A.; Wu, J.; and Amodei, D. 2020.
\newblock Scaling laws for neural language models.
\newblock \emph{arXiv preprint arXiv:2001.08361}.

\bibitem[{Klest(2020)}]{Klest:2020sdb}
Klest, H. 2020.
\newblock {Overview and design of the sPHENIX TPC}.
\newblock \emph{J. Phys. Conf. Ser.}, 1498: 012025.

\bibitem[{Kolodiazhnyi et~al.(2024)Kolodiazhnyi, Vorontsova, Konushin, and
  Rukhovich}]{kolodiazhnyi2024oneformer3d}
Kolodiazhnyi, M.; Vorontsova, A.; Konushin, A.; and Rukhovich, D. 2024.
\newblock Oneformer3d: One transformer for unified point cloud segmentation.
\newblock In \emph{Proceedings of the IEEE/CVF Conference on Computer Vision
  and Pattern Recognition}, 20943--20953.

\bibitem[{Li et~al.(2024)Li, Hu, Wang, Li, Fan, King, Song, and Li}]{Li2024}
Li, Q.; Hu, Z.; Wang, Y.; Li, L.; Fan, Y.; King, I.; Song, L.; and Li, Y. 2024.
\newblock Progress and Opportunities of Foundation Models in Bioinformatics.
\newblock \emph{Briefings in Bioinformatics}.
\newblock Survey of foundation models in bioinformatics.

\bibitem[{Loshchilov and Hutter(2017)}]{adamw}
Loshchilov, I.; and Hutter, F. 2017.
\newblock Decoupled weight decay regularization.
\newblock \emph{arXiv preprint arXiv:1711.05101}.

\bibitem[{Ma et~al.(2024)Ma, Jiang, Cheng, and Xu}]{ma2024harnessing}
Ma, Q.; Jiang, Y.; Cheng, H.; and Xu, D. 2024.
\newblock Harnessing the deep learning power of foundation models in
  single-cell omics.
\newblock \emph{Nature Reviews Molecular Cell Biology}, 25(8): 593--594.

\bibitem[{Mildenhall et~al.(2021)Mildenhall, Srinivasan, Tancik, Barron,
  Ramamoorthi, and Ng}]{nerf}
Mildenhall, B.; Srinivasan, P.~P.; Tancik, M.; Barron, J.~T.; Ramamoorthi, R.;
  and Ng, R. 2021.
\newblock Nerf: Representing scenes as neural radiance fields for view
  synthesis.
\newblock \emph{Communications of the ACM}, 65(1): 99--106.

\bibitem[{Moskowitz(2023)}]{moskowitz_tiny_2023}
Moskowitz, C. 2023.
\newblock Tiny Bubbles of Primordial Soup Re‑create Early Universe.
\newblock \emph{Scientific American}, 328(3).
\newblock Accessed: 2025‑07‑19.

\bibitem[{Nguyen et~al.(2023)Nguyen, Brandstetter, Kapoor, Gupta, and
  Grover}]{climax}
Nguyen, T.; Brandstetter, J.; Kapoor, A.; Gupta, J.~K.; and Grover, A. 2023.
\newblock ClimaX: a25 foundation model for weather and climate.
\newblock In \emph{Proceedings of the 40th International Conference on Machine
  Learning}, ICML'23. JMLR.org.

\bibitem[{Novak et~al.(2018)Novak, Bahri, Abolafia, Pennington, and
  Sohl-Dickstein}]{novak2018sensitivity}
Novak, R.; Bahri, Y.; Abolafia, D.~A.; Pennington, J.; and Sohl-Dickstein, J.
  2018.
\newblock Sensitivity and Generalization in Neural Networks: an Empirical
  Study.
\newblock In \emph{International Conference on Learning Representations}.

\bibitem[{Osborn et~al.(2021)Osborn, Frawley, Huang, Lee, Da~Costa, Peters,
  Pinkenburg, Roland, and Yu}]{Osborn:2021zlr}
Osborn, J.~D.; Frawley, A.~D.; Huang, J.; Lee, S.; Da~Costa, H.~P.; Peters, M.;
  Pinkenburg, C.; Roland, C.; and Yu, H. 2021.
\newblock {Implementation of ACTS into sPHENIX Track Reconstruction}.
\newblock \emph{Comput. Softw. Big Sci.}, 5(1): 23.

\bibitem[{Paganini(2018)}]{atlas_bjet}
Paganini, M. 2018.
\newblock {Machine Learning Algorithms for $b$-Jet Tagging at the ATLAS
  Experiment}.
\newblock \emph{J. Phys. Conf. Ser.}, 1085(4): 042031.

\bibitem[{Pyzer-Knapp et~al.(2025{\natexlab{a}})Pyzer-Knapp, Manica, Staar,
  Morin, Ruch, Laino, Smith, and Curioni}]{pyzer2025foundation}
Pyzer-Knapp, E.~O.; Manica, M.; Staar, P.; Morin, L.; Ruch, P.; Laino, T.;
  Smith, J.~R.; and Curioni, A. 2025{\natexlab{a}}.
\newblock Foundation models for materials discovery--current state and future
  directions.
\newblock \emph{Npj Computational Materials}, 11(1): 61.

\bibitem[{Pyzer-Knapp et~al.(2025{\natexlab{b}})Pyzer-Knapp, Manica, Staar,
  Morin, Ruch, Laino, Smith, and Curioni}]{PyzerKnapp2025}
Pyzer-Knapp, E.~O.; Manica, M.; Staar, P.; Morin, L.; Ruch, P.; Laino, T.;
  Smith, J.~R.; and Curioni, A. 2025{\natexlab{b}}.
\newblock Foundation models for materials discovery – current state and
  future directions.
\newblock \emph{npj Computational Materials}, 11: 15.

\bibitem[{Qasim et~al.(2019)Qasim, Kieseler, Iiyama, and
  Pierini}]{qasim2019learning}
Qasim, S.~R.; Kieseler, J.; Iiyama, Y.; and Pierini, M. 2019.
\newblock Learning representations of irregular particle-detector geometry with
  distance-weighted graph networks.
\newblock \emph{The European Physical Journal C}, 79(7): 1--11.

\bibitem[{Rusch, Bronstein, and Mishra(2023)}]{rusch2023survey}
Rusch, T.~K.; Bronstein, M.~M.; and Mishra, S. 2023.
\newblock A survey on oversmoothing in graph neural networks.
\newblock \emph{arXiv preprint arXiv:2303.10993}.

\bibitem[{Sj\"ostrand et~al.(2015)Sj\"ostrand, Ask, Christiansen, Corke, Desai,
  Ilten, Mrenna, Prestel, Rasmussen, and Skands}]{m:pythia8}
Sj\"ostrand, T.; Ask, S.; Christiansen, J.~R.; Corke, R.; Desai, N.; Ilten, P.;
  Mrenna, S.; Prestel, S.; Rasmussen, C.~O.; and Skands, P.~Z. 2015.
\newblock {An introduction to PYTHIA 8.2}.
\newblock \emph{Comput. Phys. Commun.}, 191: 159--177.

\bibitem[{{sPHENIX Collaboration}(2025{\natexlab{a}})}]{sphenix_acts}
{sPHENIX Collaboration}. 2025{\natexlab{a}}.
\newblock {acts}: A common tracking software toolkit for sPHENIX, tagged at
  \texttt{33fc284f238a24405bcd6c2de3260f370d6f8403}.
\newblock
  \url{https://github.com/sPHENIX-Collaboration/acts/tree/33fc284f238a24405bcd6c2de3260f370d6f8403}.
\newblock Accessed August 3, 2025; part of the official sPHENIX build
  environment.

\bibitem[{{sPHENIX Collaboration}(2025{\natexlab{b}})}]{sphenix_calibrations}
{sPHENIX Collaboration}. 2025{\natexlab{b}}.
\newblock {calibrations}: Calibration code for sPHENIX, tagged at
  \texttt{a3e66e69635514813ee3e20bf18b2bd59787b503}.
\newblock
  \url{https://github.com/sPHENIX-Collaboration/calibrations/tree/a3e66e69635514813ee3e20bf18b2bd59787b503}.
\newblock Accessed August 3, 2025; part of the official sPHENIX build
  environment.

\bibitem[{{sPHENIX Collaboration}(2025{\natexlab{c}})}]{sphenix_coresoftware}
{sPHENIX Collaboration}. 2025{\natexlab{c}}.
\newblock {coresoftware}: Core simulation and reconstruction software for the
  sPHENIX experiment, tagged at
  \texttt{b849eba5c2cf8ada510d036aa9b9499cb31f0513}.
\newblock
  \url{https://github.com/sPHENIX-Collaboration/coresoftware/tree/b849eba5c2cf8ada510d036aa9b9499cb31f0513}.
\newblock Accessed August 3, 2025; used in the official sPHENIX build,
  including downstream emulation and reconstruction.

\bibitem[{{sPHENIX Collaboration}(2025{\natexlab{d}})}]{sphenix_macros}
{sPHENIX Collaboration}. 2025{\natexlab{d}}.
\newblock {macros}: Analysis and utility macros for sPHENIX, tagged at
  \texttt{661f781db23352a3fa72055ce7dbf1a0ee1c2167}.
\newblock
  \url{https://github.com/sPHENIX-Collaboration/macros/tree/661f781db23352a3fa72055ce7dbf1a0ee1c2167}.
\newblock Accessed August 3, 2025; part of the official sPHENIX build
  environment.

\bibitem[{Stroud et~al.(2024)Stroud, Duckett, Hart, Pond, Rettie, Facini, and
  Scanlon}]{rw:trackingmaskformer}
Stroud, S.~V.; Duckett, P.; Hart, M.; Pond, N.; Rettie, S.; Facini, G.; and
  Scanlon, T. 2024.
\newblock Transformers for Charged Particle Track Reconstruction in High Energy
  Physics.
\newblock arXiv:2411.07149.

\bibitem[{Vankadara et~al.(2024)Vankadara, Xu, Haas, and Cevher}]{mumamba}
Vankadara, L.~C.; Xu, J.; Haas, M.; and Cevher, V. 2024.
\newblock On feature learning in structured state space models.
\newblock In \emph{The Thirty-eighth Annual Conference on Neural Information
  Processing Systems}.

\bibitem[{Vaswani et~al.(2017)Vaswani, Shazeer, Parmar, Uszkoreit, Jones,
  Gomez, Kaiser, and Polosukhin}]{rw4}
Vaswani, A.; Shazeer, N.; Parmar, N.; Uszkoreit, J.; Jones, L.; Gomez, A.~N.;
  Kaiser, {\L}.; and Polosukhin, I. 2017.
\newblock Attention is all you need.
\newblock \emph{Advances in neural information processing systems}, 30.

\bibitem[{Wang et~al.(2023)Wang, Dai, Chen, Huang, Li, Zhu, Hu, and
  et~al.}]{Wang2023}
Wang, W.; Dai, J.; Chen, Z.; Huang, Z.; Li, Z.; Zhu, X.; Hu, X.; and et~al.
  2023.
\newblock InternImage: Exploring Large-Scale Vision Foundation Models With
  Deformable Convolutions.
\newblock In \emph{CVPR 2023}, 3591--3600.

\bibitem[{Yang et~al.(2022)Yang, Hu, Babuschkin, Sidor, Liu, Farhi, Ryder,
  Pachocki, Chen, and Gao}]{mufive}
Yang, G.; Hu, E.~J.; Babuschkin, I.; Sidor, S.; Liu, X.; Farhi, D.; Ryder, N.;
  Pachocki, J.; Chen, W.; and Gao, J. 2022.
\newblock Tensor programs v: Tuning large neural networks via zero-shot
  hyperparameter transfer.
\newblock \emph{arXiv preprint arXiv:2203.03466}.

\bibitem[{Zhou et~al.(2023)Zhou, Chia, Wagner, Ayhan, Williamson, Struyven,
  Liu, Xu, Lozano, Woodward-Court et~al.}]{zhou2023foundation}
Zhou, Y.; Chia, M.~A.; Wagner, S.~K.; Ayhan, M.~S.; Williamson, D.~J.;
  Struyven, R.~R.; Liu, T.; Xu, M.; Lozano, M.~G.; Woodward-Court, P.; et~al.
  2023.
\newblock A foundation model for generalizable disease detection from retinal
  images.
\newblock \emph{Nature}, 622(7981): 156--163.

\end{thebibliography}

\end{document}